\documentclass[10pt,journal,compsoc]{IEEEtran}
\usepackage{amsmath,amsfonts}
\usepackage{amssymb}
\usepackage{algpseudocode}
\algnewcommand{\To}{\textbf{to}}  
\algrenewcommand\algorithmicrequire{\textbf{Input:}}
\algrenewcommand\algorithmicensure{\textbf{Output:}}
\usepackage{float}
\usepackage{algorithm}
\usepackage{array}
\usepackage[caption=false,font=normalsize,labelfont=sf,textfont=sf]{subfig}
\usepackage{textcomp}
\usepackage{stfloats}
\usepackage{url}
\usepackage{verbatim}
\usepackage{graphicx}
\usepackage{multirow}
\usepackage{booktabs}
\usepackage{siunitx}
\usepackage{enumitem}
\usepackage{soul}
\usepackage{mathtools}
\usepackage{makecell}
\usepackage{adjustbox}

\usepackage{cuted}
\usepackage{caption}
\usepackage{afterpage}

\ifCLASSOPTIONcompsoc
  \usepackage[nocompress]{cite}
\else
  \usepackage{cite}
\fi

\begin{document}
%
\title{Spatial-Temporal Synergy: Balancing Change and Invariance in Text-Driven 3D Human Motion Editing}
%
%
%
\author{Shaohui Lin, Zhenwu Shi, Jingyu Gong$^{*}$, Jiao Xie, Yu Zhou, Baochang Zhang, Lizhuang Ma and Chia-Wen Lin~\IEEEmembership{Fellow,~IEEE}
\IEEEcompsocitemizethanks{\IEEEcompsocthanksitem This work is supported by the National Natural Science Foundation of China (NO. 62572193, 62502159), Natural Science Foundation of Shanghai (NO. 25ZR1402135), Natural Science Foundation of Chongqing (NO. CSTB2025NSCQ-GPX0445), ``Chen Guang'' project supported by Shanghai Municipal Education Commission and Shanghai Education Development Foundation (NO. 25CGA22), China Postdoctoral Science Foundation (NO. 2024M760930), the Open Research Fund of the Key Laboratory of Advanced Theory and Application in Statistics and Data Science, Ministry of Education, and the Fundamental Research Funds for the Central Universities, Open Project Program of the State Key Laboratory of CAD\&CG (Grant No. A2501), Zhejiang University.
\IEEEcompsocthanksitem Shaohui Lin, Jingyu Gong and Lizhuang Ma are with the School of Computer Science and Technology, East China Normal University, Shanghai 200062, China. Lizhuang Ma is also with the School of Computer Science and Engineering, Shanghai Jiao Tong University, Shanghai, China. \protect\\
E-mail: \{jygong,lzma,shlin\}@cs.ecnu.edu.cn
\IEEEcompsocthanksitem Zhenwu Shi is with the Shanghai Institute of Artificial Intelligence for Education, East China Normal University, Shanghai 200062, China. \protect\\
E-mail: zw.shi@foxmail.com
\IEEEcompsocthanksitem Yu Zhou and Jiao Xie are with the School of Statistics, East China Normal University, Shanghai 200062, China. \protect\\
E-mail: \{zyu, jxie\}@stat.ecnu.edu.cn, 
\IEEEcompsocthanksitem Baochang Zhang is with the School of Artificial Intelligence, Beihang University, Beijing 100191, China, also with the Hangzhou Research Institute, Beihang University, Hangzhou 310051, China, and also with the Zhongguancun Laboratory, Beijing 100081, China. \protect\\
E-mail: bczhang@buaa.edu.cn
\IEEEcompsocthanksitem Chia-Wen Lin is the Department of Electrical Engineering, National Tsing Hua University, Hsinchu 300044, Taiwan. \protect\\
E-mail: cwlin@ee.nthu.edu.tw
\IEEEcompsocthanksitem (*Corresponding authors: Jingyu Gong.)
}}

\markboth{SUBMISSION TO IEEE TRANSACTIONS ON PATTERN ANALYSIS AND MACHINE INTELLIGENCE}%
{Shi \MakeLowercase{\textit{et al.}}: Spatial-Temporal Synergy for Text-Driven 3D Human Motion Editing}

\IEEEtitleabstractindextext{
\begin{abstract}
Text-driven human motion editing aims to modify existing motion sequences according to natural language instructions while maintaining the structural consistency of the original motion. Existing diffusion-based approaches struggle to balance text-responsive ``change'' and inertial ``invariance''. They often rely on coarse spatial constraints and rigid uniform time assumptions, leading to spatial motion distortions and the destruction of intrinsic physical rhythms during variable-length editing. To handle these challenges, we propose \textbf{C}hange and \textbf{I}nvariance \textbf{M}otion \textbf{E}diting (CIME), a unified framework that comprehensively decouples change and invariance into spatial pose and temporal rhythm dimensions. For spatial poses, our method integrates an omni-supervised positive-negative learning mechanism comprising hierarchical retrospective feature supervision, subtle motion preservation, and triplet-based semantic alignment. For temporal rhythms, we introduce the Riemannian Non-uniform Integral Manifold Mapping (RNIMM) module, which achieves high-fidelity reproduction of physical beats in the edited text via kinematics-aware non-uniform timestamps. Extensive experiments on the MotionFix and STANCE Adjustment datasets demonstrate that CIME achieves state-of-the-art performance in editing alignment and structural fidelity, validating the effectiveness of our unified architecture. Our source codes and models have been released at: github.com/ZhenwuShi/CIME.git
\end{abstract}

\begin{IEEEkeywords}
Motion Editing, Human Motion Synthesis, Diffusion Models, Omni-supervised Positive-negative Learning, Riemannian Integral Geometry 
\end{IEEEkeywords}}

\maketitle

\IEEEdisplaynontitleabstractindextext

%
\IEEEpeerreviewmaketitle

\begin{figure}
  \centering
\includegraphics[width=\linewidth,keepaspectratio]{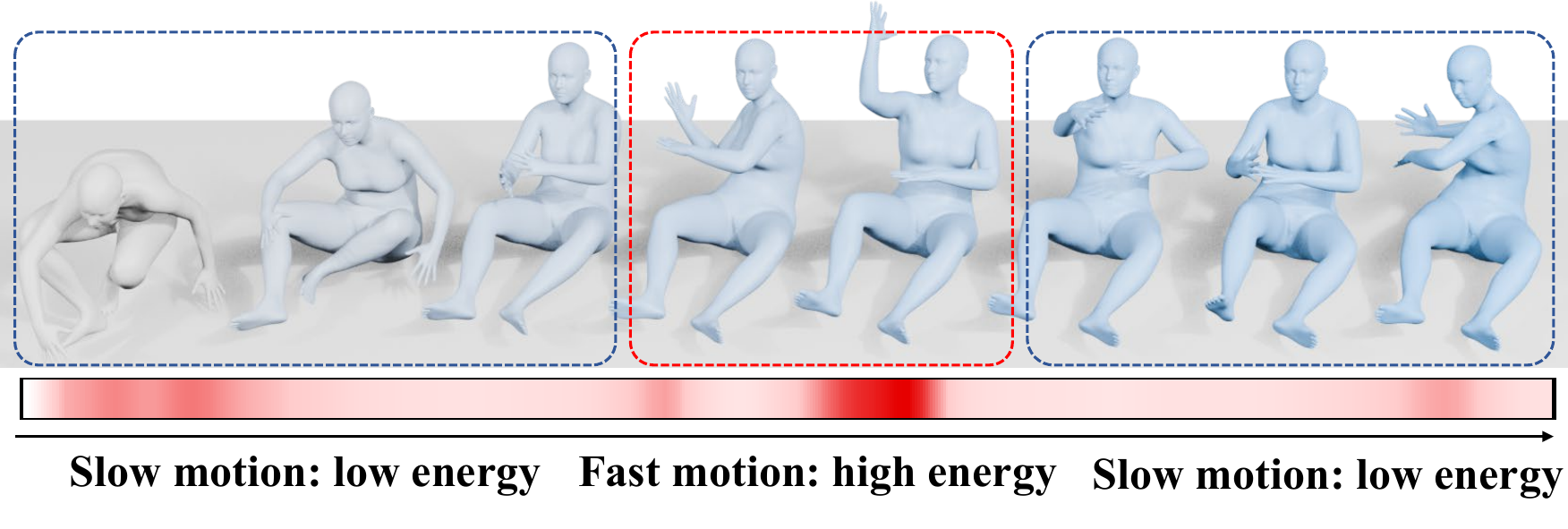}
\captionof{figure}{Display information on fast motion and slow motion.}
  \label{fig:cover2}
\end{figure}

\section{Introduction}
Text-driven 3D human motion editing has emerged as a foundational technology for intuitive animation generation, virtual reality interaction, and digital human behavior simulation. Early motion synthesis primarily focused on autoregressive pose forecasting \cite{lyu2021learning,lyu20223d,aliakbarian2020stochastic,yan2018mt}. By bridging the semantic gap between human language and 3D kinematics, modern editing technology allows users to execute fine-grained modifications on existing motion sequences simply by providing free-form natural language instructions. This capability not only significantly reduces the manual labor required by traditional animators but also enables highly efficient reuse and personalization of massive motion capture assets. Compared to generating entirely new kinematic sequences from scratch, the motion editing task aims to maximally preserve the structural properties and stylistic nuances of the original movements. Consequently, this technology demonstrates immense practical value across diverse multimedia applications, high-fidelity game development, and complex virtual reality ecosystems.

With the rapid advancement of deep generative learning, Denoising Diffusion Probabilistic Models (DDPMs) \cite{ho2020denoising,sohl2015deep} have established themselves as the dominant paradigm in human motion synthesis \cite{tevethuman}. Building upon the solid foundations of standard text-to-motion generation, researchers have progressively moved toward more flexible, interactive, and controllable motion editing tasks. Recently, pioneering frameworks such as MotionFix \cite{athanasiou2024motionfix} have introduced comprehensive text-guided benchmarks and standardized triplet datasets, driving the community toward true open-vocabulary motion manipulation rather than relying on isolated action tags. On top of these benchmarks, state-of-the-art approaches like SimMotionEdit \cite{li2025simmotionedit} have incorporated explicit motion similarity prediction as an auxiliary target to constrain the generation process. Although these models achieve certain performance, they take a relatively simplified view of the temporal alignment between the source and target sequences. Moreover, how to perfectly balance text-responsive ``change'' and inertial ``invariance'' remains an unsolved challenge, which requires meticulous synergy across both the spatial pose and temporal rhythm dimensions. 
In this work, we formally formulate this core problem as \textbf{C}hange and \textbf{I}nvariance \textbf{M}otion \textbf{E}diting (CIME). The essence of CIME is to handle the editing task from two complementary perspectives, namely spatial pose and temporal rhythm, such that text‑driven modifications can be accurately applied while the original motion structure is faithfully maintained.

\begin{figure}
  \centering
\includegraphics[width=\linewidth,keepaspectratio]{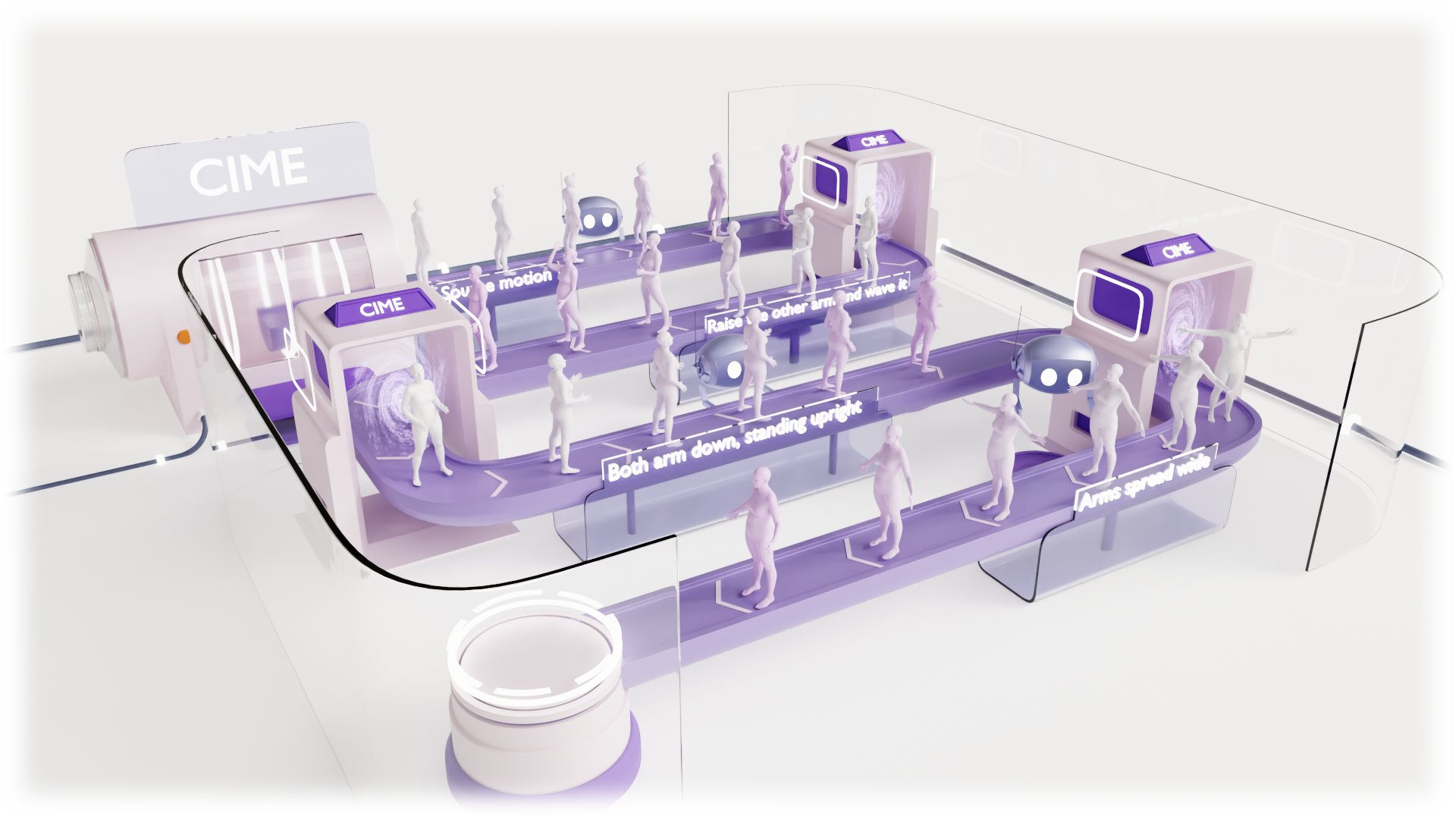}
  \captionof{figure}{\textbf{CIME} is a unified framework for text-driven 3D human motion editing. Given a source motion and a natural language instruction, CIME edits the source motion to produce the desired target motion while balancing change and invariance across both spatial pose and temporal rhythm dimensions.}
  \label{fig:cover}
\end{figure} 

At the spatial pose level, existing generative constraints are typically restricted to coarse-grained global conditions or a single terminal gradient feedback from the final network layer. Although several text-conditioned frameworks \cite{kim2023flame,chen2023executing} support editing, they often lose track of original local features during target motion generation. Without continuous intermediate constraints, such as a preservation factor $m \in [0,1]$, models executing local pose changes are easily dominated by global text instructions. This causes the network to completely overwrite the original motion details. Consequently, these methods fail to smoothly maintain structural invariance in unaffected body regions. This specific limitation directly motivates our design of a continuous spatial constraint. More critically, a severe temporal mapping dilemma arises during variable-length editing. Because the input source motion and the generated target one often differ in total frame count, they inherently lack a direct, frame-to-frame correspondence. Traditional motion diffusion models and temporal generation frameworks \cite{petrovich2022temos,zhang2024motiondiffuse} almost universally default to a flat and uniform Euclidean time assumption. This assumption is strictly bound to rigid, equidistant frame indices. Unfortunately, when textual instructions demand macroscopic changes in speed or duration, this rigid time structure forces a uniform stretching or compressing of the timeline. This will definitely undermine the rhythmic modifications induced by the textual instructions for motion editing.

Our core motivation stems from the critical necessity to break this uniform time assumption and establish a flexible, dynamic mapping mechanism that naturally accommodates variable-length editing. Since the source and target sequences are not strictly one-to-one, we inherently require a non-uniform mapping strategy to automatically synchronize different kinematic timelines rather than relying on fixed frame indices. Furthermore, pure time-domain feature learning struggles to master subtle velocity fluctuations and complex rhythm variations, such as fast actions or sudden localized pauses (as visually illustrated in Fig.~\ref{fig:cover2}). To capture these minute temporal deviations and explicitly learn the complex transitions of motion speed, we need to rethink the temporal architecture and model the timing axis based on the intrinsic geometric properties of the motion itself. By mapping the motion progress to a non-uniform timeline, the network can naturally preserve the intrinsic physical dynamics without being distorted by macroscopic duration discrepancies.

To implement the aforementioned concepts, we propose a unified framework termed CIME (as illustrated in Fig.~\ref{fig:cover}), which comprehensively decouples "change and invariance" into two core dimensions: temporal rhythm and spatial pose. For the spatial pose, the framework integrates a comprehensive omni-supervised positive-negative learning mechanism to regulate spatial structural integrity. For the temporal rhythm, we introduce the Riemannian Non-uniform Integral Manifold Mapping (RNIMM) module to explicitly model the timing axis.

A preliminary version of this work was presented as OmniME\cite{Shi_2026_CVPR} in CVPR 2026. The earlier version primarily focused on the spatial dimension, relying solely on the omni-supervised mechanism for pose editing. This journal manuscript fundamentally extends the preliminary work into a ``spatial-temporal synergy'' paradigm. Crucially, the newly proposed RNIMM module leverages Riemannian integral geometry to compute kinematics-aware non-uniform timestamps via arc-length parameterization, replacing the standard uniform time assumption and successfully preserving intrinsic physical beats against sequence length variations.

We conduct extensive evaluations on the MotionFix and STANCE Adjustment datasets. To thoroughly validate the proposed extensions, we significantly expand our experimental scope by introducing fine-grained diagnostic analyses and zero-shot cross-dataset generalization tests. The results systematically demonstrate that CIME not only achieves state-of-the-art retrieval accuracy but also exhibits exceptional robustness across different domains.

Our contributions are summarized as follows:
\begin{itemize}
    \item We propose CIME, a novel and unified framework for text-driven 3D human motion editing that explicitly balances text-responsive change and inertial invariance across both spatial pose and temporal rhythm dimensions.
    \item We design a robust omni-supervised positive-negative learning mechanism that exerts tight constraints on pose change and invariance across feature, motion, and semantic levels, significantly enhancing spatial precision and preventing unwanted joint distortions.
    \item We propose the Riemannian Non-uniform Integral Manifold Mapping (RNIMM) module, which aims to tackle the temporal mapping dilemma under variable-length editing by utilizing kinematics-aware non-uniform timestamps and preserves intrinsic physical rhythms.
\end{itemize}

\section{Related Work}

\subsection{Human Motion Generation}

Driven by large-scale kinematic datasets~\cite{guo2022generating,mahmood2019amass,punnakkal2021babel,shahroudy2016ntu}, early motion synthesis primarily focused on autoregressive pose forecasting~\cite{aliakbarian2020stochastic,yan2018mt,lyu2021learning,lyu20223d}. To bridge the semantic gap, subsequent deep learning approaches introduced natural language and discrete action categories as conditioning signals to enhance behavioral alignment~\cite{bie2022hit,harvey2020robust,hong2022avatarclip,lin2018human,petrovich2022temos,tevet2022motionclip,wang2020learning,zhong2023attt2m,fan2024textual}. 

Recently, Denoising Diffusion Probabilistic Models (DDPMs)~\cite{ho2020denoising,sohl2015deep} have fundamentally shifted motion synthesis by formulating it as a gradual denoising process, achieving state-of-the-art performance in text-driven generation~\cite{chen2023executing,dabral2023mofusion,dai2024motionlcm,liang2024omg,liang2024intergen,ren2023insactor,tevethuman,xu2023interdiff,zhang2023remodiffuse,zhang2024motiondiffuse}. Moving beyond isolated action tags, recent frameworks leverage sentence-level datasets~\cite{guo2022generating,lin2023motion,punnakkal2021babel} to interpret open-ended linguistic descriptions~\cite{shafirhuman,tevethuman,zhang2024motiondiffuse,wan2023diffusionphase,xie2023omnicontrol}. To ensure sequence fidelity, methods like Lagrangian Motion Fields~\cite{yang2025lagrangian} enhance temporal dynamics for long-term generation, while multimodal frameworks like MotionLLM~\cite{chen2025motionllm} leverage the reasoning capabilities of Large Language Models to achieve deeper alignment between complex linguistic instructions and fine-grained behaviors.

Building upon these foundations, diffusion-based methods have naturally extended to spatial-temporal editing tasks like inbetweening and inpainting~\cite{tevethuman,shafirhuman}. Concurrently, advanced approaches such as FineMoGen~\cite{zhang2023finemogen} harness LLMs to facilitate localized motion editing. This progression, alongside new benchmarks for open-vocabulary generation~\cite{wang2025you}, underscores the ongoing demand for semantic-aware, highly generalizable motion editing systems. While these approaches excel at open-ended synthesis, our framework specifically targets the structural preservation of source motions, enabling semantic-aware editing through a continuous spatial constraint mechanism.

\subsection{Human Motion Editing}

Early motion editing heavily relied on spatial and temporal constraints~\cite{gleicher1997motion,lee1999hierarchical,gleicher2001motion} for tasks like trajectory alteration~\cite{kim2009synchronized}, skeletal retargeting~\cite{aberman2020skeleton}, and style transfer~\cite{unuma1995fourier}. Deep learning subsequently advanced stylistic translation~\cite{aberman2020skeleton,yin2023dance}, localized pose refinement~\cite{oreshkinprotores}, and temporal in-betweening~\cite{qin2022motion,shafirhuman,tseng2023edge}. However, achieving fine-grained editing strictly through text remained a formidable challenge.

Diffusion models have recently revitalized this domain. Researchers have explored mask-based spatial inpainting~\cite{kim2023flame} and latent query substitution~\cite{raab2024monkey}. While several text-conditioned frameworks~\cite{pinyoanuntapong2024mmm,zhang2024motiondiffuse} support editing, they often hard-freeze unedited joints, which severely disrupts kinematic fluidity. Alternative compositional frameworks~\cite{athanasiou2022teach,shi2024interactive,athanasiou2023sinc,petrovich2022temos,petrovich2024multi} offer smoother transitions but are bottlenecked by heavy manual annotation requirements.

To enable more intuitive interaction, recent pipelines like PoseFix~\cite{delmas2023posefix} utilize free-form natural language. Concurrently, LLM-empowered techniques~\cite{zhang2023finemogen,huang2024controllable} manipulate motions at textual or latent levels. For standardized evaluation, MotionFix~\cite{athanasiou2024motionfix} contributed a comprehensive text-guided editing benchmark. Advanced systems~\cite{zhang2023finemogen,goel2024iterative,huang2024controllable} further refined semantic control, though they inherently suffer from dense annotation burdens. Additionally, recent efforts explicitly predict joint-wise motion differences via cross-axis feature fusion for precise text-based editing~\cite{han2026cross}. Motivated by these benchmarks and datasets~\cite{athanasiou2024motionfix}, we propose a framework to balance change and invariance across both spatial poses and temporal rhythms.

\subsection{Generative Model Supervision}
Effective supervision mechanisms are crucial for synthesizing realistic and semantically faithful outputs. In deep generative models, perceptual constraints~\cite{johnson2016perceptual} and adversarial feature matching~\cite{salimans2016improved} frequently preserve high-fidelity structural details. Concurrently, metric-based optimization, such as triplet and contrastive regularization~\cite{schroff2015facenet,chen2020simple}, effectively organizes the underlying latent spaces.

For human motion editing, supervision must simultaneously address cross-modal semantic alignment and kinematic continuity. MotionCLIP~\cite{tevet2022motionclip} achieves semantic alignment using pretrained contrastive encoders. To standardize text-guided editing, MotionFix~\cite{athanasiou2024motionfix} provides a diffusion baseline optimized on annotated triplets comprising source motions, target behaviors, and textual instructions. Meanwhile, diffusion frameworks~\cite{tevethuman,shafirhuman} incorporate explicit temporal-consistency penalties to ensure smooth and natural kinematic transitions during editing tasks.

Recently, researchers have enriched supervisory signals via novel data curation and auxiliary objectives. Dynamic augmentations like MotionCutMix~\cite{jiang2025dynamic} expand the training distribution, while frameworks like SimMotionEdit~\cite{li2025simmotionedit} use auxiliary similarity-prediction targets to implicitly constrain generation. Furthermore, advanced systems like FineMoGen~\cite{zhang2023finemogen} leverage LLMs to construct fine-grained supervisory signals for localized spatial control. Different from existing discrete or auxiliary supervisory objectives, our framework employs a continuous geometric supervision approach that explicitly regulates the balance between temporal rhythm invariance and local pose adaptability.

\subsection{Riemannian Geometry and Temporal Manifolds}
Riemannian geometry and manifold learning have been widely applied in human motion analysis. However, recent state-of-the-art research has predominantly focused on modeling the spatial constraints of skeletal poses. For instance, FlashMo \cite{zhang2026flashmo} proposes geometric interpolation on Lie group manifolds to ensure the consistency of joint rotations; NRMF \cite{yu2026geometric} constructs neural distance fields on Riemannian motion manifolds to explicitly model the physical priors of pose, velocity, and acceleration; and RMG \cite{miao2026riemannian} further explores motion generation on non-Euclidean product manifolds via Riemannian flow matching.

Despite significant advancements in spatial geometric constraints, existing mainstream motion diffusion models \cite{zhang2024motiondiffuse} and temporal generation frameworks \cite{wang2025timotion} almost universally default to a flat and uniform Euclidean time assumption for temporal modeling, relying entirely on discrete and equidistant integer frame indices for feature progression. Departing from these Euclidean paradigms, we pioneer the introduction of Riemannian geodesics directly into the modeling of the time axis itself. By mapping the cumulative physical work of the motion to geodesic arc length, we construct a non-uniform temporal architecture that locks the invariant intrinsic physical beats against macroscopic sequence length changes.

\section{Methods}
In the following parts, we will first establish the preliminaries of motion representation and formulate our editing objective in Section 3.1. Then, we will give an overview of our comprehensive motion editing framework in Section 3.2. Next, we will introduce the Semantic Alignment via Triplet Learning in Section 3.3, followed by the RNIMM module for temporal alignment and rhythm preservation in Section 3.4. Finally, we will present two additional supervision mechanisms that co-optimize the diffusion backbone, namely the Retrospective Feature Supervision in Section 3.5, and the Motion Preservation during editing in Section 3.6.

\subsection{Preliminary}
\textbf{Motion Representation.} 
Let $\mathbf{X} = [\mathbf{x}_0, \mathbf{x}_1, \dots, \mathbf{x}_F]$ denote a continuous sequence of human motion frames, where each $\mathbf{x}_i$ captures the comprehensive full-body pose at a specific temporal step. Adhering to the standard configuration established by MotionFix~\cite{athanasiou2024motionfix}, we encode the frame-level feature into a 207-dimensional vector:
\begin{equation}
\mathbf{x}_i = [\mathbf{v}_i, \mathbf{o}_i, \mathbf{r}_i, \mathbf{p}_i] \in \mathbb{R}^{207}.
\end{equation}
Within this formulation, the individual components explicitly correspond to the global root velocity, the global spatial orientation, the local joint rotations, and the localized 3D joint positions, respectively.

\textbf{Motion Editing.}
The primary objective of text-guided motion editing is to synthesize a targeted kinematic sequence by leveraging a given source motion $\mathbf{X}$ alongside a natural language instruction $\mathbf{L}$. This generative mapping can be abstracted as $\mathbf{M} = \mathcal{G}(\mathbf{X}, \mathbf{L})$, with $\mathcal{G}$ serving as the conditional synthesis network. 

\textbf{Editing Objective.}
First, the synthesized motion must accurately match the specific requirements of the textual instruction. Second, it preserves the inherent characteristics and unmodified regions of the raw source motion. Third, the entire motion sequence retains rigorous kinematic continuity and spatiotemporal coherence, ensuring smooth transition between motion frames without jarring artifacts.

Following the standard Denoising Diffusion Probabilistic Models~\cite{ho2020denoising, sohl2015deep}, the primary diffusion loss is defined as:
\begin{equation}
\mathcal{L}_{\mathrm{diff}} = \mathbb{E}_{t, \mathbf{x}_0, \epsilon, \mathbf{c}} \left\| \epsilon - \epsilon_\theta \left( \mathbf{x}_t, \, t, \, \mathbf{c} \right) \right\|_2^2,
\end{equation}
where $\mathbf{x}_t = \sqrt{\bar{\alpha}_t}\,\mathbf{x}_0 + \sqrt{1 - \bar{\alpha}_t}\,\epsilon$ is the noisy target motion, $\mathbf{c}$ denotes the combined conditioning from the text instruction and source motion, and $\epsilon_\theta$ is the denoising network.

\subsection{Overview}

\begin{figure*}[t]
    \centering
    \includegraphics[width=\linewidth]{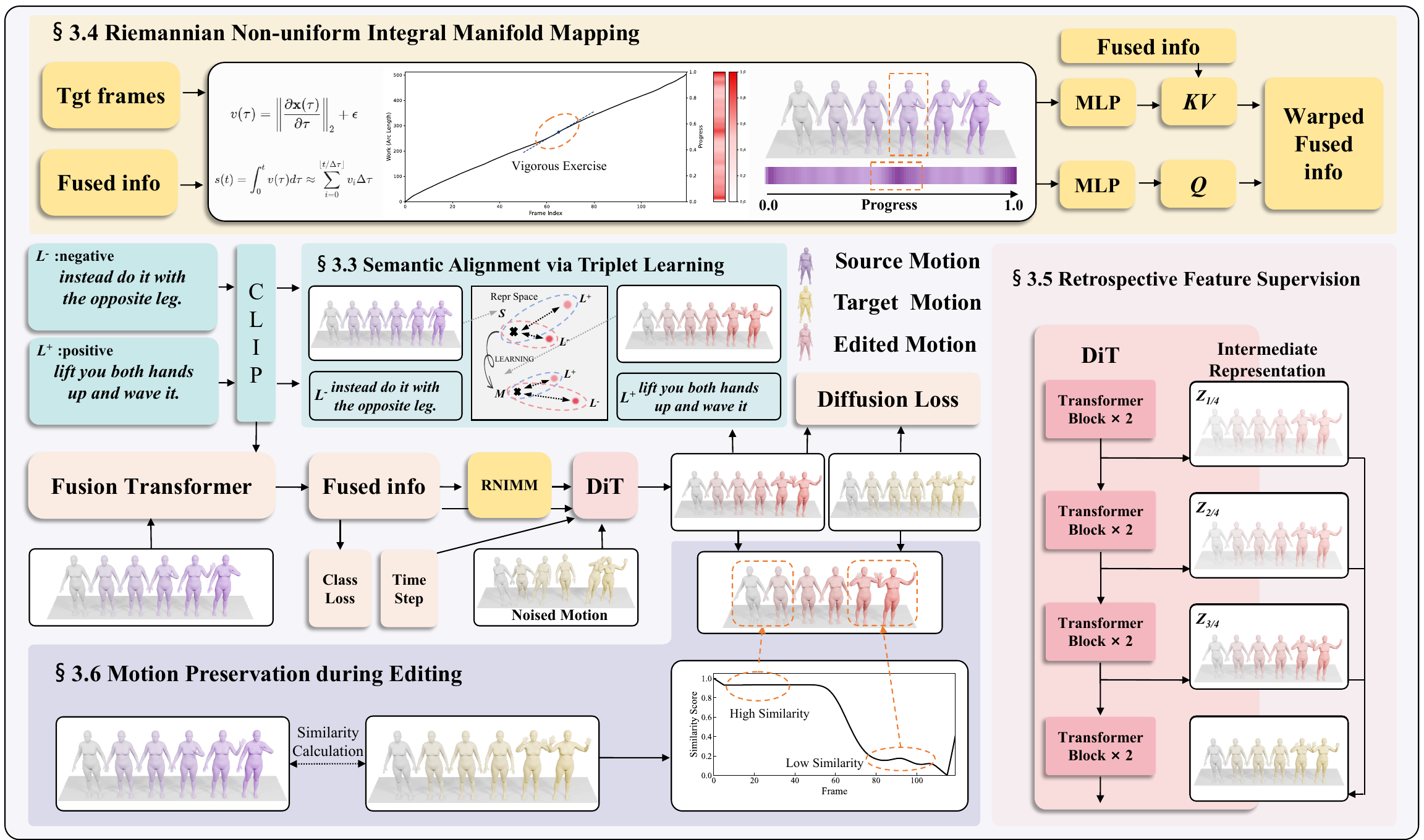}  
    \caption{\textbf{Overview of CIME}: First, the input text and source motion are projected into a shared semantic space via a CLIP encoder and aggregated by a Fusion Transformer. Next, the RNIMM module aligns these text-infused features with the target temporal dimension and injects them into the noise target. Finally, the composite features are processed by a Diffusion Transformer. This backbone is co-optimized by three auxiliary supervision mechanisms to achieve an optimal balance between precise editing and structure preservation, ultimately yielding the target motion.}
    \label{fig:overview}
\end{figure*}

As depicted in Fig.~\ref{fig:overview}, our framework balances the core philosophy of ``change and invariance''. First, a pre-trained CLIP~\cite{radford2021learning} encoder projects both textual and kinematic modalities into a shared semantic space. A Fusion Transformer then aggregates them to generate fused features. Crucially, these fused features encapsulate the textual semantics that fundamentally drive the motion modifications. Alongside the noise target's frame length, they are processed by the Riemannian Non-uniform Integral Manifold Mapping module. This module aligns the features to the target timeline and injects them into the noise target, resolving the conflict between macroscopic length variations and intrinsic rhythm preservation.

The composite features are subsequently routed into a Diffusion Transformer~\cite{peebles2023scalable}. To ensure high-quality synthesis, this backbone is co-optimized by three distinct mechanisms. First, a retrospective strategy explicitly constrains intermediate representations. Second, a dynamic preservation module maintains foundational postures in unedited regions. Third, a contrastive triplet objective strictly fortifies semantic alignment. Ultimately, the synergy of these modules achieves a perfect equilibrium between precise text-driven changes and high-fidelity structural invariance.

\subsection{Semantic Alignment via Triplet Learning}

To achieve strict semantic synchronization between edited motions and text prompts, a contrastive triplet objective operating directly on the terminal representations of the transformer backbone is integrated. Assuming $\mathbf{h}^{(L)} \in \mathbb{R}^{B \times T \times D}$ acts as the hidden feature map extracted from the final network layer, we collapse the temporal dimension through global average pooling to derive a compact motion embedding $\mathbf{z}_m \in \mathbb{R}^{B \times D}$:
\begin{equation}
    \mathbf{z}_m = \frac{1}{T} \sum_{t=1}^{T} \mathbf{h}^{(L)}_t, \label{eq:motion_embed}
\end{equation}
where $T$ indicates the temporal duration of the sequence.

Subsequently, we designate $\mathbf{z}_p \in \mathbb{R}^{B \times D}$ as the positive linguistic condition intrinsically matching the target behavior, while $\mathbf{z}_n \in \mathbb{R}^{B \times D}$ serves as a randomly sampled negative textual feature. The optimization landscape driving this semantic alignment is governed by the following triplet formulation:
\begin{equation}
    \mathcal{L}_{\text{triplet}} = \frac{1}{B} \sum_{i=1}^{B} \Big[ \|\mathbf{z}_m^i - \mathbf{z}_p^i\|_2^2 - \|\mathbf{z}_m^i - \mathbf{z}_n^i\|_2^2 + \alpha \Big]_+. \label{eq:triplet}
\end{equation}
Here, the hinge function $[\cdot]_+$ outputs the maximum between its argument and zero, while $\alpha$ represents the minimum geometric margin enforced between positive and negative pairs, which we empirically configure to $0.2$ throughout our experiments. This contrastive regularization strictly compels the generated motion manifold to gravitate towards the correct textual semantics while actively repelling mismatched instructions, thereby guaranteeing high fidelity to the user's intent.

\subsection{Riemannian Non-uniform Integral Manifold Mapping}

\textbf{Kinematics-Aware Riemannian Geodesic Construction.}
Existing motion editing models generally treat human motion as a sequence of discrete frames with uniform time mappings. However, real human movements naturally involve varying intensities of physical momentum, such as explosive fast punches or localized static pauses. In addition, textual instructions also impose rhythmic adjustments on a subset of the motions therein. To strictly preserve the intrinsic physical dynamics and to capture both the changes and invariances dictated by the textual instructions, we break the assumption of uniform time mapping. Instead, we construct non-uniform geodesic timestamps on a Riemannian manifold based on kinematic features.
Inspired by the arc-length parameterization in Riemannian geometry, 
we model a motion sequence as a continuous trajectory 
$\gamma(t)$ evolving on an underlying latent motion manifold. 
The intrinsic length of this trajectory can be formulated as:

\begin{equation}
L(\gamma)
=
\int_{0}^{T}
\sqrt{
\dot{\gamma}(t)^{T}
g(\gamma(t))
\dot{\gamma}(t)
}
dt ,
\label{eq:riemannian_length}
\end{equation}
where $g(\gamma(t))$ denotes the Riemannian metric tensor that measures 
the local geometric structure of the latent manifold. 
However, practical motion representations are inherently discrete, 
where each sequence consists of finite latent motion features 
$F=\{f_0,f_1,\dots,f_{T-1}\}$.
Therefore, the continuous geodesic length is approximated by accumulating 
the local displacements between adjacent frames:

\begin{equation}
s_i
=
\sum_{k=0}^{i}\tilde{v}_k ,
\label{eq:arc_length_discrete}
\end{equation}
where $s_i$ represents the accumulated intrinsic arc length up to the 
$i$-th frame, and $\tilde{v}_k$ denotes the regularized local variation.
Specifically, the instantaneous motion variation between neighboring 
latent states is computed as:

\begin{equation}
v_i
=
\left\|
f_i-f_{i-1}
\right\|_2 ,
\label{eq:local_velocity}
\end{equation}
where $v_i$ measures the kinematic change magnitude between two consecutive 
motion states. To avoid zero-length intervals caused by static motion 
segments, we introduce a small positive constant $\epsilon$:

\begin{equation}
\tilde{v}_i
=
v_i+\epsilon .
\label{eq:regularized_velocity}
\end{equation}

\begin{figure}
  \centering
\includegraphics[width=\linewidth,keepaspectratio]{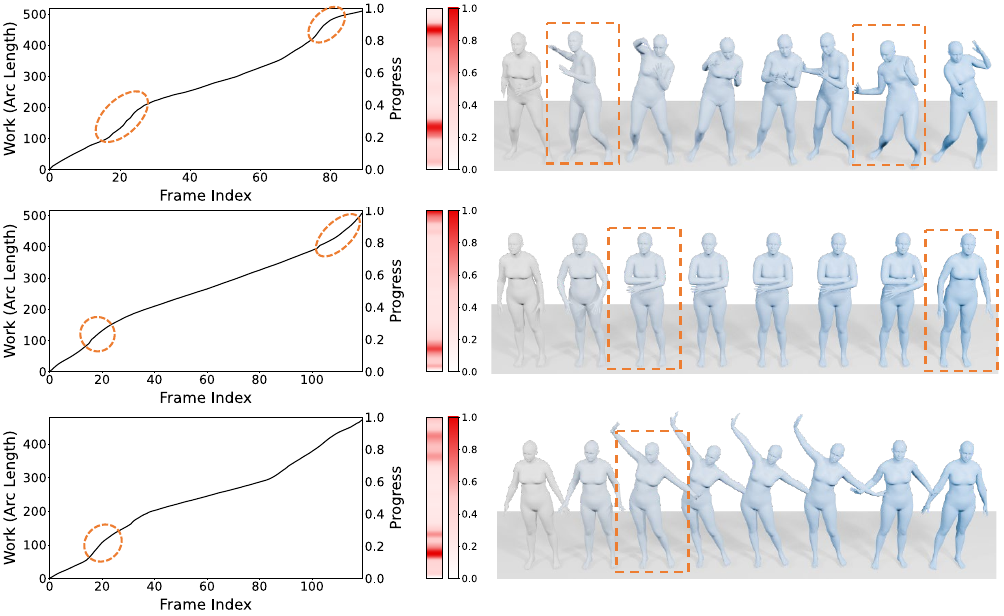}
  \captionof{figure}{Illustration of arc length on Riemannian manifolds and encoding on non-uniform Riemannian manifolds. The orange boxes indicate regions of intense motion, and the red bars reflect higher instantaneous kinematic intensity.}
  \label{fig:RME}
\end{figure}

The intrinsic temporal coordinate of each frame is obtained by 
normalizing the accumulated arc length:

\begin{equation}
t_i^{\mathrm{intrinsic}}
=
\frac{s_i}{s_{T-1}} .
\label{eq:intrinsic_timestamp}
\end{equation}
Finally, we divide the cumulative arc length of the current frame by the total arc length of the last frame. This yields the normalized physical rhythm progress $t_{src}^{(i)}$:
\begin{equation}
t_{src}^{(i)} = \frac{s_i}{s_{T_{src}-1}}, \quad t_{src}^{(i)} \in [0, 1]
\end{equation}
Through this mechanism, each source frame receives a non-uniform temporal coordinate in the normalized arc-length parameter space that reflects its kinematic intensity: as illustrated in Fig.~\ref{fig:RME} frames in high-velocity regions are distributed more sparsely along the [0, 1] progress axis, while those in quasi-static regions are clustered more densely. This non-uniform temporal parameterization causes the downstream cross-sequence alignment to adaptively allocate greater temporal resolution toward dynamically rich regions, faithfully preserving the underlying kinematic rhythm even as the sequence duration varies, and providing a robust geometric baseline for subsequent cross-length motion editing.

After establishing the geodesic time for the source features, we construct the temporal mapping for the target sequence. Since the target motion is generated from pure noise via diffusion, it lacks any initial physical structure. Unlike the non-uniform progress of the source, the target sequence is strictly modeled as a Euclidean time manifold with a constant progression rate.

Given the target length $T_{tgt}$ determined by textual instructions, we perform uniform linear sampling within the closed interval $[0, 1]$ to generate the target timestamps:
\begin{equation}
t_{tgt}^{(j)} = \frac{j}{T_{tgt} - 1}, \quad j \in \{0, 1, \dots, T_{tgt}-1\}
\end{equation}

Regardless of how the target duration scales, this uniform ruler—through the optimal transport alignment described below—adaptively re-samples features from the non-uniform Riemannian manifold. Consequently, it decouples macroscopic sequence length changes from the preservation of localized physical beats.

\textbf{Rhythm Mapping.}
To inject the constructed one-dimensional non-uniform timestamps and overcome spectral bias, we introduce a high-frequency implicit neural representation. We apply a Fourier positional encoding $\gamma(t)$ with logarithmically decaying frequencies, followed by a multi-layer perceptron to yield high-dimensional temporal features $E_t \in \mathbb{R}^D$. To compute the first-order semantic cost $M_{cost}$, these temporal features are explicitly added to the spatial source motion features to construct the keys and values. This mapping is indispensable, as it amplifies minute temporal deviations from the Riemannian geodesic that standard linear projections would obscure.

For downstream alignment, we propose a log-domain Fused Gromov-Wasserstein mechanism. To preserve the source kinematic rhythm, we define the intra-manifold topological distances as:
\begin{equation}
C_{src} = (t_{src} \ominus t_{src})^2, \quad C_{tgt} = (t_{tgt} \ominus t_{tgt})^2
\end{equation}
Combined with the semantic cost $M_{cost}$, the optimal transport plan $P$ is stably solved via log-space Sinkhorn iterations, effectively preventing numerical overflow. Finally, we resample the source features using $P$. This seamlessly aligns the non-uniform Riemannian manifold with the uniform Euclidean timeline, yielding the aligned representation for the subsequent diffusion process.

\subsection{Retrospective Feature Supervision}

Building upon SimMotionEdit~\cite{li2025simmotionedit}, we incorporate a retrospective feature supervision mechanism~\cite{tan2023positive,xiang2023retro} to enhance the optimization stability of the generative process. Our denoising backbone is instantiated as a Diffusion Transformer~\cite{peebles2023scalable} comprising $8$ sequential blocks. Breaking away from the conventional paradigm that relies exclusively on the terminal layer for gradient feedback, we establish multi-level intermediate constraints. Specifically, we append lightweight linear projection heads to the hidden states of the $2$-nd, $4$-th, and $6$-th intermediate blocks.

Formally, assume $\mathbf{h}^{(l)} \in \mathbb{R}^{B \times T \times D}$ represents the internal feature map extracted from the $l$-th block, where the dimensions indicate the batch size, sequence length, and hidden channels, respectively. The auxiliary projection layer $f^{(l)}(\cdot)$ transforms these hidden features directly back into the target kinematic space:
\begin{equation}
    \hat{\mathbf{x}}^{(l)} = f^{(l)}(\mathbf{h}^{(l)}), \quad l \in \{2,4,6\}, \label{eq:retro_pred}
\end{equation}
Here, $\hat{\mathbf{x}}^{(l)} \in \mathbb{R}^{B \times T \times J}$ signifies the recovered motion sequence, with $J$ denoting the dimensions of the motion features.

To enforce structural consistency, we calculate the mean squared error between these intermediate predictions and the actual ground-truth motion $\mathbf{x}$:
\begin{equation}
    \mathcal{L}^{(l)} = \frac{1}{BTJ} \sum_{b=1}^{B} \sum_{t=1}^{T} \| \hat{\mathbf{x}}^{(l)}_{b,t} - \mathbf{x}_{b,t} \|_2^2. \label{eq:retro_mse}
\end{equation}

The overall retrospective objective is subsequently formulated as a weighted summation of these intermediate penalties:
\begin{equation}
    \mathcal{L}_{\text{retro}} = \sum_{l \in \{2,4,6\}} \lambda_l \, \mathcal{L}^{(l)}, \label{eq:retro_total}
\end{equation}
The coefficient $\lambda_l$ modulates the intensity of the supervisory signals injected at different architectural depths. In execution, the reconstruction errors derived from the designated intermediate blocks are averaged prior to integration with the primary reconstruction loss originating from the final $8$-th block. This hierarchical aggregation guarantees that the terminal network layer retains its strict priority in guiding the motion generation. Ultimately, this deep supervision strategy explicitly forces the internal feature representations to converge toward the target data distribution at an earlier stage, thereby substantially mitigating training volatility during motion editing.

\subsection{Motion Preservation during Editing}

Building upon the foundational principles of SimMotionEdit~\cite{li2025simmotionedit}, we first evaluate frame-level kinematic similarities to explicitly isolate the temporal regions that necessitate structural preservation during the modification process. Assuming a source sequence $\mathbf{x} = \{x_i\}_{i=1}^T$ and its corresponding synthesized counterpart $\mathbf{m} = \{m_j\}_{j=1}^T$, this similarity metric is derived through a three-stage progressive pipeline.

\textbf{Raw Similarity.}
Initially, we quantify the structural proximity across both rotational and positional spaces by employing a local sliding window of size $W$. The rotational discrepancy is mathematically defined as:
\begin{equation}
S_i^{Rr} = -\min_{|i-j|\leq W} d_r(x_i, m_j),
\end{equation}
where $d_r(\cdot)$ acts as a standard distance metric, such as the Euclidean norm, applied within the orientation manifold. Let $S_i^{Rr}$ and $S_i^{Rl}$ represent the computed frame-wise similarities for joint rotations and local positions, respectively. The unified raw similarity score is subsequently formulated via a weighted linear combination:
\begin{equation}
S_i^{R} = w_1 S_i^{Rr} + w_2 S_i^{Rl}.
\end{equation}

\textbf{Normalized Similarity.}
To guarantee scale-invariant evaluations across diverse action categories, these raw metric values undergo a stringent normalization bounding them within the closed interval $[0,1]$. This transformation yields a continuous temporal similarity profile, which effectively accentuates the frames exhibiting substantial divergence between the source and target behaviors while preserving uniform numerical magnitudes across all processed sequences.

\textbf{Motion Signal-to-Noise Filtering.}
To suppress irrelevant fluctuations and distill robust supervisory signals, a bespoke Motion Signal-to-Noise Ratio is calculated. This metric is formulated as:
\begin{equation}
\text{MotionSNR} = \frac{\sum_{x \in T^R} x}{\sum_{x \in B^R} x},
\end{equation}
Here, $T^R$ and $B^R$ denote the subsets containing the top-$\kappa$ and bottom-$\kappa$ frames, respectively, as ranked by their temporal similarity scores. We selectively retain data pairs exhibiting elevated MotionSNR values, as they more accurately distinguish the mutable editing windows from the rigid background context. Adhering to the empirical configurations established by SimMotionEdit~\cite{li2025simmotionedit} on the MotionFix~\cite{athanasiou2024motionfix} dataset, we configure the sliding window parameter $W$ alongside weighting coefficients $w_1 = 1.0$ and $w_2 = 1.0$. Specifically for the STANCE Adjustment dataset, we set $\kappa = 5$ while establishing the minimum signal-to-noise threshold at $1.5$.

\textbf{Motion Preservation.}
A pronounced MotionSNR empirically indicates that the synthesized behavior remains structurally homologous to the source motion, implying that the textual instruction demands only localized and nuanced spatial adjustments. For these specific motion pairs surpassing the predefined threshold $\tau$, we enforce a dedicated preservation loss to penalize any unwarranted deviations from the original kinematic blueprint:
\begin{equation}
\mathcal{L}_{\text{presv}} = \mathbb{I}\big(\text{MotionSNR}(\mathbf{x},\mathbf{m})>\tau\big)\cdot \frac{1}{T}\sum_{i=1}^{T}\|m_i - x_i\|_2^2.
\end{equation}
where $\mathbb{I}(\cdot)$ serves as the binary indicator function, with $T$ denoting the total sequence duration. This regularization objective strictly reinforces the structural integrity within high-confidence samples, thereby steering the network to concentrate its computational capacity exclusively on the semantic modifications dictated by the text while shielding the foundational motion architecture from degradation.

Overall, our framework establishes comprehensive supervision over the ``change and invariance'' of poses. The $\mathcal{L}_{\text{cls}}$ directly follows the definition from SimMotionEdit\cite{li2025simmotionedit}. The global loss function is thus expressed as:
\begin{equation}
\begin{aligned}
\mathcal{L}_{\text{total}} = & \ \mathcal{L}_{\text{diff}} + \lambda_{\text{cls}} \mathcal{L}_{\text{cls}} + \lambda_{\text{retro}} \mathcal{L}_{\text{retro}} \\
& + \lambda_{\text{preserve}} \mathcal{L}_{\text{presv}} + \lambda_{\text{triplet}} \mathcal{L}_{\text{triplet}}.
\end{aligned}
\end{equation}
The configurations of the loss weights and implementation details are detailed in Section 4.4.

\begin{figure*}[t]
    \centering
    \includegraphics[width=\linewidth]{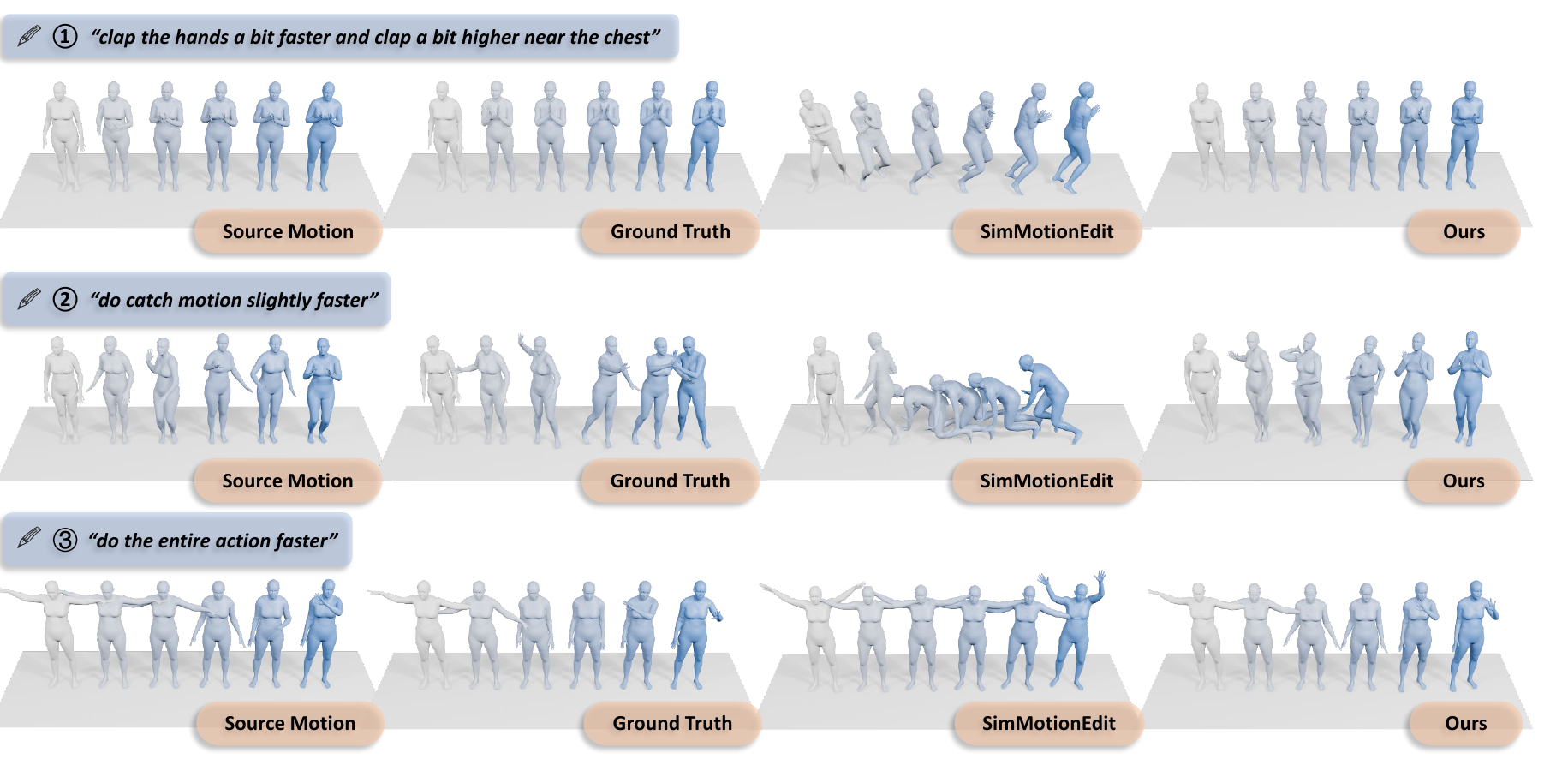}
    \captionof{figure}{\textbf{Qualitative comparison between our method and SimMotionEdit on the MotionFix dataset.} Our results surpass SimMotionEdit in terms of semantic consistency, motion smoothness, and source motion preservation.}  
    \label{fig:mot_vis}
    
    \vspace{2mm}  
    
    \includegraphics[width=\linewidth]{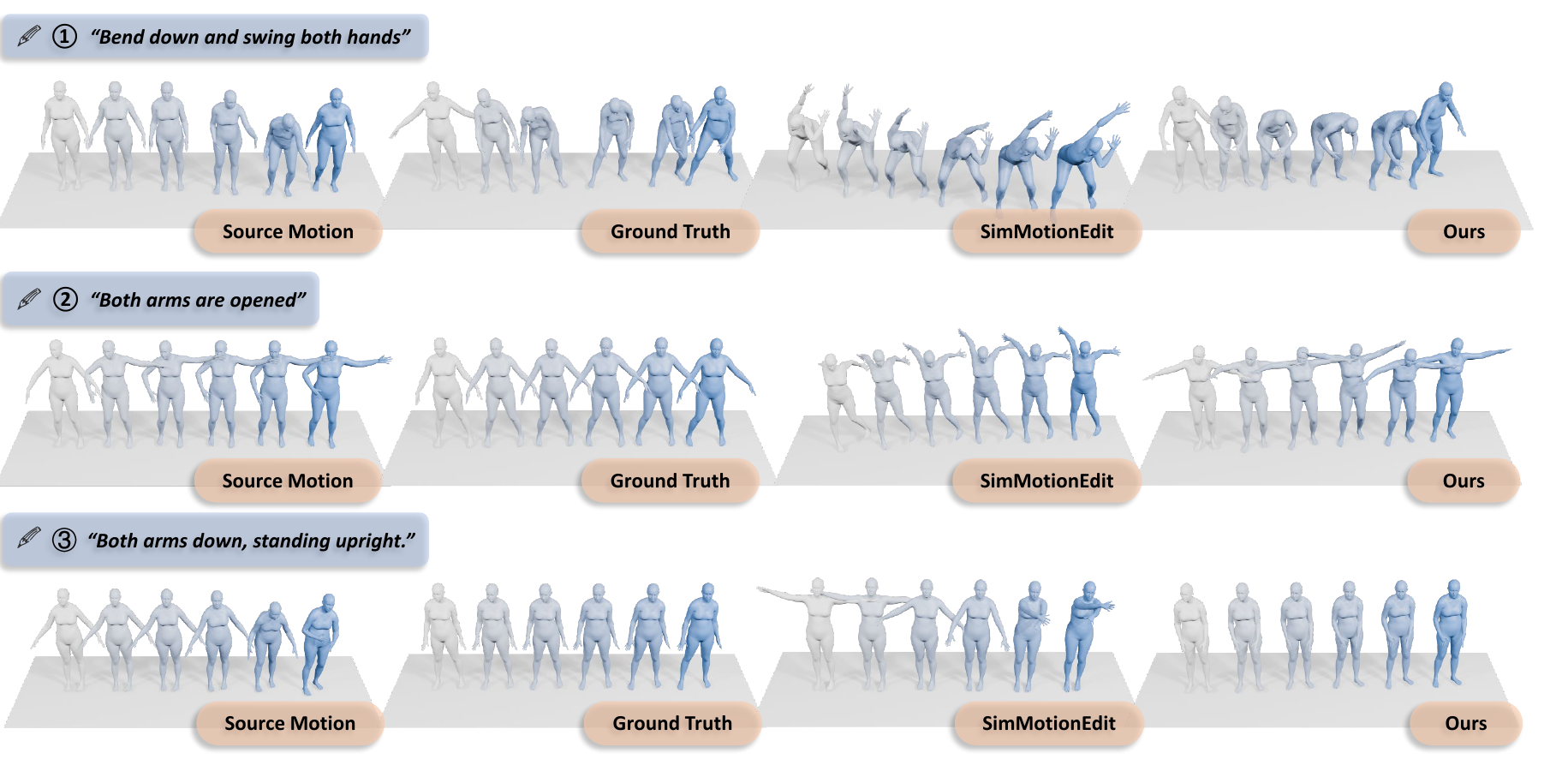}
    \captionof{figure}{\textbf{Qualitative comparison between our method and SimMotionEdit on the STANCE Adjustment dataset.} Our results surpass SimMotionEdit in terms of semantic consistency, motion smoothness, and source motion preservation.}  
    \label{fig:stance_vis}
    \vspace{-1mm}  
\end{figure*}

\section{Experiments}
In the following parts, we will first introduce the experimental datasets, evaluation metrics, competitive baselines, and implementation details in Sections 4.1, 4.2, 4.3, and 4.4, respectively. Then, we will present the quantitative analysis and qualitative results to demonstrate the superiority of our comprehensive framework in Sections 4.5 and 4.6. Next, we will provide extensive ablation studies to validate the effectiveness of our core components and distinct supervision strategies in Section 4.7. Finally, we will assess the subjective perceptual quality of our generated motions in Section 4.8, and explore the out-of-distribution robustness of our model in Section 4.9.

\begin{table*}[t]
\centering
\caption{\textbf{Comparison of retrieval performance (Generated-to-Target) on MotionFix dataset~\cite{athanasiou2024motionfix}}. In the following tables and figures, 
$R@k$ indicates recall at top-$k$ ($\uparrow$ higher is better, $\downarrow$ lower is better). * : no explicit text conditioning in the DiT stage.}
\resizebox{\textwidth}{!}{
\begin{tabular}{llcccccccc}
\toprule
\multirow{2}{*}{Method} & \multirow{2}{*}{Venue} & 
\multicolumn{4}{c}{Generated-to-Target (Batch)} & 
\multicolumn{4}{c}{Generated-to-Target (Test Set)} \\
\cmidrule(lr){3-6} \cmidrule(lr){7-10}
& & R@1$\uparrow$ & R@2$\uparrow$ & R@3$\uparrow$ & AvgR$\downarrow$
& R@1$\uparrow$ & R@2$\uparrow$ & R@3$\uparrow$ & AvgR$\downarrow$ \\
\midrule
Ground Truth        & -- & 100   & 100   & 100   & 1.00  & 64.36 & 88.75 & 95.56 & 1.74 \\
MDM \cite{tevethuman}     & ICLR'23 & 4.03  & 7.56  & 10.48 & 15.55 & 0.10  & 0.10  & 0.10  & --    \\
MDM-BP  \cite{athanasiou2024motionfix} & SIGGRAPH~Asia'24 & 39.10  & 50.09 & 54.84 & 6.46  & 8.69  & 14.71 & 18.36 & 180.99 \\
TMED \cite{athanasiou2024motionfix}  & SIGGRAPH~Asia'24 & 62.90  & 76.51 & 83.06 & 2.71  & 14.51 & 21.72 & 28.73 & 56.63 \\
MotionReFit \cite{jiang2025dynamic} & CVPR'25 & 66.33 & 80.05 & 84.98 & 2.64  & --   & --   & --   & --    \\
SimMotionEdit \cite{li2025simmotionedit} & CVPR'25 & 70.62 & 82.92 & 88.12 & 2.38  & 25.49 & 39.33 & 49.21 & 23.49 \\
SimMotionEdit * \cite{li2025simmotionedit} & CVPR'25 & 71.04 & 83.96 & 89.58 & 2.22    & 26.88    & 44.27    & 51.98    & 20.88     \\
Han et al.\cite{han2026cross} & CVPR'26 & 74.38 & \textbf{88.54} & 92.08 & 1.92  & 29.45 & 45.26 & 54.55 & 16.42 \\
OmniME \cite{Shi_2026_CVPR}  & CVPR'26 & 77.29 & \textbf{88.54} & 91.88 & 1.79  & 32.02 & 50.20 & \textbf{59.88} & 13.06 \\
\textbf{CIME(Ours)}             & -- & \textbf{78.12} & \textbf{88.54} & \textbf{92.50} & \textbf{1.73} & \textbf{33.40} & \textbf{50.29} & \textbf{59.88} & \textbf{12.24} \\
\bottomrule
\end{tabular}
}
\label{tab:retrieval_results_1}
\end{table*}

\begin{table*}[t]
\centering
\caption{\textbf{Comparison of retrieval performance (Generated-to-Target) on STANCE Adjustment dataset~\cite{jiang2025dynamic}}.}
\resizebox{\textwidth}{!}{
\begin{tabular}{l l cccc cccc}
\toprule
\multirow{2}{*}{Method} & \multirow{2}{*}{Venue} & \multicolumn{4}{c}{Generated-to-Target (Batch)} & \multicolumn{4}{c}{Generated-to-Target (Test Set)} \\
\cmidrule(lr){3-6} \cmidrule(lr){7-10}
& & R@1$\uparrow$ & R@2$\uparrow$ & R@3$\uparrow$ & AvgR$\downarrow$
  & R@1$\uparrow$ & R@2$\uparrow$ & R@3$\uparrow$ & AvgR$\downarrow$ \\
\midrule
TMED \cite{athanasiou2024motionfix} & SIGGRAPH~Asia'24 
& 29.69 & 44.01 & 52.08 & 6.97 & 11.22 & 17.86 & 25.51 & 35.56 \\

TMED w/ MCM \cite{athanasiou2024motionfix,jiang2025dynamic} & CVPR'25 
& 32.22 & 45.03 & 54.83 & 6.56 & - & - & - & - \\

SimMotionEdit * \cite{li2025simmotionedit} & CVPR'25 
& 36.46 & 48.96 & 57.81 & 5.71 & 12.76 & 23.98 & 29.59 & 29.05 \\

MotionReFit \cite{jiang2025dynamic} & CVPR'25 
& 42.45 & 56.25 & 62.76 & 5.12 & - & - & - & - \\

OmniME\cite{Shi_2026_CVPR} & CVPR'26 
& 43.75 & 56.25 & 66.15 & 4.66 
& 22.45 & 31.63 & 36.22 & 22.77 \\
\textbf{CIME} & --- & \textbf{47.92} & \textbf{61.46} & \textbf{71.35} & \textbf{4.34} & \textbf{29.59} & \textbf{36.22} & \textbf{41.33} & \textbf{22.44} \\
\bottomrule
\end{tabular}
}
\label{tab:retrieval_results_2}
\end{table*}

\section{Experimental Setups}
\textbf{Dataset.} 
To train and evaluate our framework, we utilize the MotionFix dataset~\cite{athanasiou2024motionfix}, which supplies meticulously annotated triplets comprising a source behavior, a target behavior, and a linguistic instruction. The curation of this corpus relies on the TMR model~\cite{petrovich2023tmr} to mine structurally similar kinematic pairs from extensive MoCap archives, after which human annotators articulate the precise semantic deltas between them. In total, this collection furnishes $6,730$ distinct triplets distributed across the training, validation, and testing phases.

Additionally, we employ the STANCE Adjustment dataset~\cite{jiang2025dynamic}, an auxiliary corpus constructed upon the generative foundations of the MLD architecture~\cite{chen2023executing}. During its compilation, the latent space of the generator is systematically perturbed to yield $16$ distinct kinematic variants for every base instruction. Consequently, any given pair drawn from these variants encapsulates a nuanced spatial or temporal transformation. Human experts subsequently transcribe these subtle modifications into natural language prompts, culminating in a supplementary repository of $4,411$ curated triplets.

\textbf{Evaluation metrics.} 
The fundamental imperative in text-guided kinematic editing is to ensure strict consistency between the synthesized outputs and the ground-truth target behaviors. Adhering to the standardized evaluation pipeline established by MotionFix~\cite{athanasiou2024motionfix}, we implement a motion-to-motion retrieval paradigm. In practice, this entails projecting the generated sequences into a comparable latent space utilizing the pre-trained TMR encoder~\cite{petrovich2023tmr}, followed by calculating the retrieval precision within a constrained batch of candidates. To provide a comprehensive quantitative assessment, we report the Top-1, Top-2, and Top-3 retrieval accuracies (commonly denoted as R@1, R@2, and R@3) evaluated both within a batch size of $32$ and across the entire testing distribution, alongside the average retrieval rank for in-depth benchmarking.

\textbf{Baselines.}
To rigorously evaluate our framework, we benchmark its performance against six representative state-of-the-art models: MDM~\cite{tevethuman}, MDM-BP~\cite{athanasiou2024motionfix}, TMED~\cite{athanasiou2024motionfix}, SimMotionEdit~\cite{li2025simmotionedit}, MotionReFit~\cite{jiang2025dynamic}, and a recent cross-axis fusion approach~\cite{han2026cross}. Each competitor embodies a distinct paradigm within the motion editing domain. MDM~\cite{tevethuman} demonstrates pure diffusion editing without observing source kinematics. MDM-BP~\cite{athanasiou2024motionfix} injects source constraints but lacks targeted instruction-guided optimization. TMED~\cite{athanasiou2024motionfix} provides a strong baseline by systematically integrating linguistic conditions and source behaviors. Among the latest advancements, SimMotionEdit~\cite{li2025simmotionedit} enforces semantic fidelity via similarity prediction mechanisms. MotionReFit~\cite{jiang2025dynamic} boosts generation flexibility and temporal coherence using a dedicated coordinator and dynamic augmentation. Additionally, Han et al.~\cite{han2026cross} propose a cross-axis feature fusion architecture and employ joint-wise motion difference prediction as an auxiliary task to achieve precise, localized edits. 

\textbf{Implementation details.} 
During the generative process, the diffusion trajectory spans $300$ steps governed by a cosine noise scheduler. Aligning with established practices in conditional generation~\cite{athanasiou2024motionfix,brooks2023instructpix2pix}, we enforce a two-way conditioning strategy, allocating a uniform guidance scale of $2.0$ for both the textual and source motion inputs. The semantic features of the language instructions are extracted utilizing a pre-trained CLIP ViT-L/14 encoder~\cite{radford2021learning}. Architecturally, the fusion module and the diffusion backbone are instantiated as Transformers comprising $4$ and $8$ layers, respectively. Both networks share a unified configuration of $8$ attention heads mapped to a $512$-dimensional latent space. 

Network optimization is executed via the AdamW solver~\cite{loshchilovdecoupled} configured with a learning rate of $1\times 10^{-4}$ and a batch size of $128$. For the RNIMM module, the scaling coefficient utilized to inject the rhythm-aligned source features into the noise target is explicitly set to $\lambda_{\text{RNIMM}} = 0.05$. Across all empirical setups, the global loss coefficients are rigidly maintained at $\lambda_{\text{retro}} = 1.0$ and $\lambda_{\text{triplet}} = 0.01$. Meanwhile, the motion preservation weight $\lambda_{\text{preserve}}$ is configured to $0.2$ when training on the MotionFix dataset~\cite{athanasiou2024motionfix}, and adjusted to $0.1$ for the STANCE Adjustment corpus~\cite{jiang2025dynamic}. The entire framework undergoes $1,500$ epochs of training on a single NVIDIA A6000 GPU. Computationally, this requires approximately $19$ hours to converge on the MotionFix dataset and $11$ hours on the STANCE Adjustment dataset.

\subsection{Quantitative Analysis}
We present quantitative comparisons in Tab.~\ref{tab:retrieval_results_1} and Tab.~\ref{tab:retrieval_results_2}. Our model consistently performs better than all baselines in motion-editing alignment. As expected, MDM~\cite{tevethuman} exhibits the weakest performance, highlighting that text instructions alone are insufficient to produce source-aligned edits. Incorporating GPT-based identification of editable body parts, MDM-BP~\cite{athanasiou2024motionfix} shows notable improvements over MDM. TMED~\cite{athanasiou2024motionfix}, which models the influence of text instructions on the source motion, further surpasses MDM-BP. MotionReFit~\cite{jiang2025dynamic} benefits from MCM to enrich training data, providing additional gains, while SimMotionEdit~\cite{li2025simmotionedit} leverages source-target similarity curves to design auxiliary loss functions that enhance alignment. Moreover, Han \textit{et al.}~\cite{han2026cross} propose a cross-axis feature fusion architecture and employ joint-wise motion difference prediction as an auxiliary task to achieve precise, localized edits. Although Han \textit{et al.}~\cite{han2026cross}, OmniME~\cite{Shi_2026_CVPR}, and our proposed CIME achieve an identical Batch R@2 score of 88.54, sample-level analysis reveals this is merely a numerical coincidence with non-identical retrieved subsets, and CIME ultimately demonstrates superior robustness by maintaining higher rankings for non-hit samples. 

\begin{table*}[t]
\centering
\caption{\textbf{Ablation study on the MotionFix dataset~\cite{athanasiou2024motionfix}}. Comparison of retrieval performance (Generated-to-Target).}
\resizebox{0.8\textwidth}{!}{
\begin{tabular}{c @{\hskip 4pt}c@{\hskip 4pt}c@{\hskip 4pt}c@{\hskip 4pt}c| cccccccc}
\toprule
\multirow{2}{*}{\#} & \multicolumn{4}{c|}{Components} & \multicolumn{4}{c}{Generated-to-Target (Batch)} & \multicolumn{4}{c}{Generated-to-Target (Test Set)} \\
\cmidrule(lr){2-5} \cmidrule(lr){6-9} \cmidrule(lr){10-13}
 & $\mathcal{L}_{\text{retro}}$ & $\mathcal{L}_{\text{triplet}}$ & $\mathcal{L}_{\text{pres}}$ & $\mathcal{M}_{\text{RNIMM}}$
 & R@1$\uparrow$ & R@2$\uparrow$ & R@3$\uparrow$ & AvgR$\downarrow$
 & R@1$\uparrow$ & R@2$\uparrow$ & R@3$\uparrow$ & AvgR$\downarrow$ \\
\midrule
1 &  &  &  &  & 71.04 & 83.96 & 89.58 & 2.22 & 26.88 & 44.27 & 51.98 & 20.88 \\ 
2 & \checkmark &  &  &  & 72.71 & 85.62 & 88.75 & 2.09 & 30.63 & 45.26 & 54.35 & 18.62 \\ 
3 &  & \checkmark &  &  & 73.54 & 85.00 & 88.54 & 2.04 & 28.26 & 42.49 & 51.78 & 17.99 \\ 
4 &  &  & \checkmark &  & 75.62 & 87.50 & 90.42 & 1.88 & 30.24 & 45.65 & 56.92 & 15.58 \\ 
5 &  &  &  & \checkmark & 73.54 & 87.92 & 91.04 & 1.88 & 29.64 & 44.27 & 55.34 & 15.71 \\ 
6 & \checkmark & \checkmark & \checkmark & \checkmark & \textbf{78.12} & \textbf{88.54} & \textbf{92.50} & \textbf{1.73} & \textbf{33.40} & \textbf{50.29} & \textbf{59.88} & \textbf{12.24} \\ 
\bottomrule
\end{tabular}
}
\label{tab:ablation_results}
\end{table*}

\begin{table*}[t]
\centering
\caption{\textbf{Analysis of Negative Text Sampling Strategies.} The evaluation is conducted on the MotionFix dataset~\cite{athanasiou2024motionfix}.}
\resizebox{0.75\textwidth}{!}{
\begin{tabular}{lcccccccc}
\toprule
\multirow{2}{*}{Method} & \multicolumn{4}{c}{Generated-to-Target (Batch)} & \multicolumn{4}{c}{Generated-to-Target (Test Set)} \\
\cmidrule(lr){2-5} \cmidrule(lr){6-9}
& R@1$\uparrow$ & R@2$\uparrow$ & R@3$\uparrow$ & AvgR$\downarrow$
& R@1$\uparrow$ & R@2$\uparrow$ & R@3$\uparrow$ & AvgR$\downarrow$ \\
\midrule
Baseline & 71.04 & 83.96 & \textbf{89.58} & 2.22 & 26.88 & \textbf{44.27} & \textbf{51.98} & 20.88 \\
infoNCE~\cite{radford2021learning}   & 68.33   & 83.12   & 86.04  & 2.38    & 25.10 & 38.14 & 48.62 & 24.14 \\
K-means        & 70.21   & 82.50   & 87.92  & 2.35    & 27.47 & 42.29 & 50.79 & 21.06 \\
\textbf{Triplet(Ours)}  & \textbf{73.54} & \textbf{85.00} & 88.54 & \textbf{2.04}  & \textbf{28.26} & 42.49 & 51.78 & \textbf{17.99} \\
\bottomrule
\end{tabular}
}
\label{tab:ablation_triplet}
\end{table*}

\subsection{Qualitative Results}
We present a visual comparison of our method against SimMotionEdit~\cite{li2025simmotionedit}. Fig.~\ref{fig:mot_vis} and~\ref{fig:stance_vis} show examples from the MotionFix~\cite{athanasiou2024motionfix} and STANCE Adjustment~\cite{jiang2025dynamic} datasets.

Fig.~\ref{fig:mot_vis} shows examples from MotionFix~\cite{athanasiou2024motionfix}. In Example 1, although SimMotionEdit generates roughly correct movements, it introduces an unnatural self-rotation. In contrast, our method produces stable and realistic motions. In Example 2, the motion generated by SimMotionEdit is completely unrelated to the textual instruction, whereas our approach aligns perfectly with the target semantics. In Example 3, SimMotionEdit struggles with incorrect temporal rhythm and spatial amplitude, while our method ensures accurate physical dynamics and appropriate motion scales.

Fig.~\ref{fig:stance_vis} shows examples from STANCE Adjustment~\cite{jiang2025dynamic}. In Example 1, SimMotionEdit exhibits an abnormal arm-raising behavior that is absent from both the source motion and the instruction, whereas our method maintains proper structural integrity. In Example 2, given the instruction ``Both arms are opened'', SimMotionEdit produces an exaggerated amplitude leading to highly unnatural physical deformities. Our method, however, achieves a natural and anatomically consistent adjustment. Finally, in Example 3, SimMotionEdit fails to follow the textual instruction and merely reconstructs the source motion, while our method accurately applies the intended modifications.

Overall, our approach outperforms SimMotionEdit~\cite{li2025simmotionedit} in semantic understanding, physical plausibility, and motion fidelity, producing motions that are more natural, coherent, and faithful to textual instructions.

\begin{table}[t!]
\centering
\caption{Comparison with Curriculum Scheduling based on Generated-to-Target (Batch).}
\resizebox{\columnwidth}{!}{
\begin{tabular}{lcccc}
\toprule
Method & R@1$\uparrow$ & R@2$\uparrow$ & R@3$\uparrow$ & AvgR$\downarrow$ \\
\midrule
Curriculum Scheduling    & 72.92  & 84.79 & \textbf{90.21} & 2.07 \\
Ours  & \textbf{73.54} & \textbf{85.00} & 88.54 & \textbf{2.04} \\
\bottomrule
\end{tabular}
}
\label{tab:Curriculum_ablation}
\end{table}


\begin{table*}[t]
\centering
\caption{Trained on MotionFix and tested on STANCE adjustment. ${\dag}$ indicates our re-implementation.}
\resizebox{0.8\textwidth}{!}{
\begin{tabular}{llcccccccc}
\toprule
\multirow{2}{*}{Method} & \multirow{2}{*}{Venue} & 
\multicolumn{4}{c}{Generated-to-Target (Batch)} & 
\multicolumn{4}{c}{Generated-to-Target (Test Set)} \\
\cmidrule(lr){3-6} \cmidrule(lr){7-10}
& & R@1$\uparrow$ & R@2$\uparrow$ & R@3$\uparrow$ & AvgR$\downarrow$
& R@1$\uparrow$ & R@2$\uparrow$ & R@3$\uparrow$ & AvgR$\downarrow$ \\
\midrule
SimMotionEdit*$^{\dag}$~\cite{li2025simmotionedit} & CVPR'25 & 21.43 & 33.04 & 42.41 & 8.80 & 7.42 & 11.35 & 14.41 & 58.67 \\
OmniME \cite{Shi_2026_CVPR} & CVPR'26 & 22.40 & 35.94 & 47.40 & 7.44 & 6.63 & 10.71 & 15.82 & 41.23 \\
\textbf{CIME (Ours)} & --- & \textbf{29.69} & \textbf{43.23} & \textbf{54.69} & \textbf{6.36} & \textbf{10.20} & \textbf{18.37} & \textbf{23.47} & \textbf{34.74} \\
\bottomrule
\end{tabular}
}
\label{tab:method_comparison}
\end{table*}

\subsection{Ablation Study}
Tab.~\ref{tab:ablation_results} presents the results of our ablation study on the MotionFix~\cite{athanasiou2024motionfix} dataset. We evaluate the individual contributions of the retrospective loss ($\mathcal{L}_{\text{retro}}$), triplet loss ($\mathcal{L}_{\text{triplet}}$), preservation loss ($\mathcal{L}_{\text{preserve}}$), and the RNIMM module. As shown, removing any of these components leads to a consistent performance drop. Specifically, $\mathcal{L}_{\text{retro}}$ improves structural stability through multi-level feature constraints. $\mathcal{L}_{\text{preserve}}$ ensures the original kinematics of unedited frames are maintained. The RNIMM module effectively aligns temporal rhythms across varying sequence lengths. Finally, $\mathcal{L}_{\text{triplet}}$ enforces strict semantic alignment between the generated motions and the textual instructions. The full model, which combines all these mechanisms, achieves the best overall performance. This demonstrates the effectiveness and complementarity of our proposed modules.

We further compare different strategies for constructing negative texts, with the results presented in Table~\ref{tab:ablation_triplet}. 
(1) The Triplet Loss (Random Negatives) randomly samples negative texts, enabling the model to learn from a broad and diverse range of semantic contrasts, thereby improving generalization. 
(2) The Cluster-based Triplet Loss employs K-means to partition all texts into three clusters, selecting negative samples from the same cluster as the positive text. Although this design aims to capture fine-grained distinctions within similar semantics, it limits the model's exposure to diverse negative samples, leading to insufficient generalization. 
(3) The InfoNCE Loss~\cite{radford2021learning}, following the CLIP paradigm, treats all other samples within a batch as negatives. However, since motion editing instructions often repeat across samples, semantically identical instructions may appear as false negatives, confusing the contrastive objective and degrading performance.

Furthermore, we investigate a Curriculum Scheduling approach, which dynamically transitions from easy (random) negatives to hard (in-cluster) negatives during the training process. Intuitively, curriculum learning should refine semantic boundaries; however, our empirical results (Tab.~\ref{tab:Curriculum_ablation}) demonstrate that purely random negative sampling performs better and exhibits greater stability. The underlying reason is that, in the context of motion editing, in-cluster ``hard'' negatives are often structurally ambiguous. They typically share highly overlapping kinematic trajectories with the positive samples but differ only in nuanced semantics. Forcing the model to strictly repel these structurally similar samples in later training stages introduces conflicting supervision, which disrupts the continuous motion manifold and destabilizes convergence. Consequently, the random-negative triplet loss achieves the most optimal and stable balance between diversity and semantic discrimination.

\subsection{Perceptual Study}
\begin{figure}[t]
    \centering
            \includegraphics[width=\linewidth]{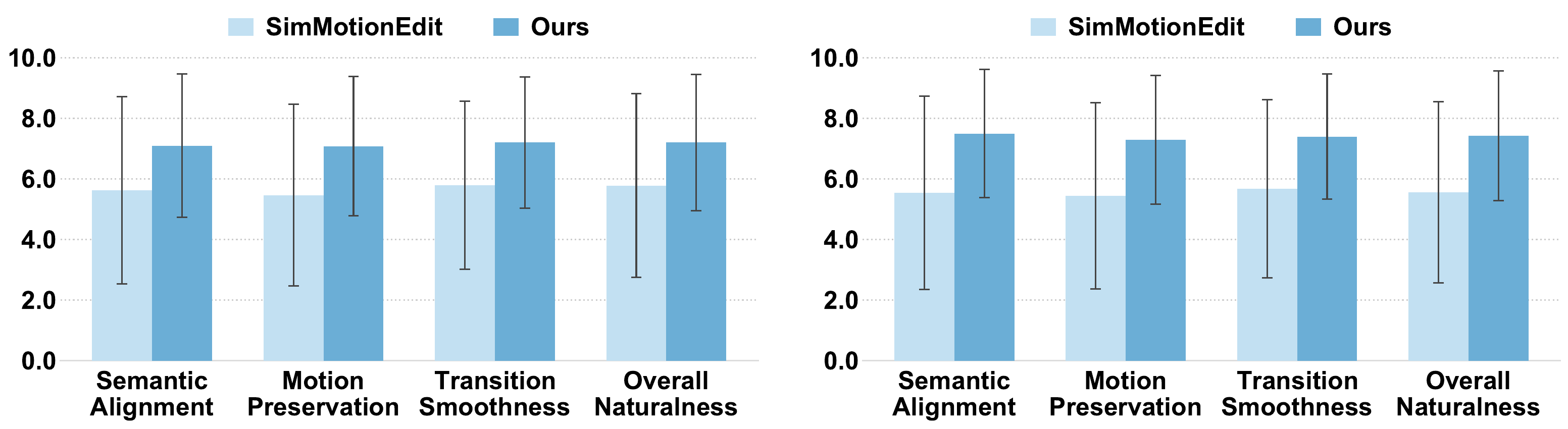}
    \vspace{-2mm}
        \caption{
        \textbf{ Subjective evaluation: CIME vs SimMotionEdit\cite{li2025simmotionedit}.} 
                The charts report average human ratings (scale 1--10) across four perceptual metrics: Semantic Alignment, Motion Preservation, Transition Smoothness, and Overall Naturalness. Results for the MotionFix~\cite{athanasiou2024motionfix} and STANCE Adjustment~\cite{jiang2025dynamic} datasets are displayed on the \textit{left} and \textit{right}, respectively. Our proposed CIME consistently outperforms the baseline across all metrics on both datasets.}
    \label{fig:userstudy}
    \vspace{-2mm}
\end{figure}

To assess the perceptual quality of our approach compared to SimMotionEdit~\cite{li2025simmotionedit}, we organized a subjective evaluation. The study involved 25 volunteers, comprising academic professors, graduate students, and industry professionals. We selected 20 representative editing cases in total. Half of these were sourced from the MotionFix~\cite{athanasiou2024motionfix} dataset, and the remaining half from the STANCE Adjustment~\cite{jiang2025dynamic} dataset. During the test, participants reviewed the original kinematics, textual prompts, ground-truth targets, and the synthesized outputs. For a fair and unbiased comparison, the evaluated models were blindly labeled as Method A and Method B.

Evaluators scored the generated sequences from 1 (worst) to 10 (best) based on four criteria: Semantic Alignment, Motion Preservation, Transition Smoothness, and Overall Naturalness. As illustrated in Fig.~\ref{fig:userstudy}, CIME achieved consistently higher scores than SimMotionEdit~\cite{li2025simmotionedit} across all evaluated dimensions. It demonstrates superior textual responsiveness and structural integrity. These subjective results validate the effectiveness of our design philosophy of change and invariance.

\subsection{Out-of-Distribution Robustness}
Although achieving optimal cross-dataset generalization typically necessitates dedicated Domain Adaptation techniques---a distinct and expansive research area designated for our future exploration---we nonetheless conduct cross-dataset benchmarking to validate the intrinsic robustness of our architecture. Specifically, the evaluated models undergo training exclusively on the MotionFix~\cite{athanasiou2024motionfix} corpus and are subsequently deployed directly onto the unseen STANCE Adjustment~\cite{jiang2025dynamic} dataset for zero-shot testing. As illustrated in Tab.~\ref{tab:method_comparison}, even in the complete absence of domain-specific fine-tuning, our proposed CIME framework consistently and significantly surpasses both SimMotionEdit~\cite{li2025simmotionedit} and the prior OmniME\cite{Shi_2026_CVPR} architecture across all evaluated retrieval metrics. This substantial performance leap empirically confirms the formidable generalization capacity of our approach. Furthermore, it emphasizes that the delicate equilibrium between structural invariance and semantic modification established by CIME is not merely overfitted to the idiosyncratic distributions of a specific training dataset, but rather profoundly captures the fundamental principles and universal physical laws governing text-driven motion editing.


\section{Conclusion}
In this paper, we propose CIME, a unified framework for text-driven 3D human motion editing. CIME effectively balances text-responsive modifications with the preservation of original motion structures. We achieve this by decoupling the editing process into spatial and temporal dimensions. For spatial poses, we introduce an omni-supervised learning mechanism to maintain structural accuracy. For temporal rhythms, we present the RNIMM module to align varying sequence lengths while preserving intrinsic physical beats. Experiments confirm that CIME achieves state-of-the-art performance across multiple benchmarks.

Despite these advancements, our work has certain limitations. Achieving optimal generalization across entirely different motion datasets remains a challenge. Specifically, our framework currently lacks cross-dataset generalization capabilities. In future work, we plan to explore dedicated Domain Adaptation techniques. This future research direction will further enhance the model's cross-dataset robustness and expand its generalizable applicability.

\bibliographystyle{IEEEtran}
\bibliography{references}

\begin{IEEEbiography}[{\includegraphics[width=1in,height=1.25in,clip,keepaspectratio]{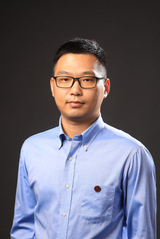}}]
{Shaohui Lin} received Ph.D. degree from Xiamen University, Xiamen, China, in 2019. He is currently a Research Professor at East China Normal University. He is the author of about 40 scientific articles at top venues, including IEEE TPAMI, TIP, CVPR, ICML, and ICLR, etc. He serves as a Area Chair at top-tier conferences, such as CVPR, ICML and ICLR, as well as  a reviewer for IEEE TPAMI, IJCV, TMM and CVPR, etc. His research interests include machine learning and computer vision.
\end{IEEEbiography}

\vspace{-2em}
\begin{IEEEbiography}[{\includegraphics[width=1in,height=1.25in,clip,keepaspectratio]{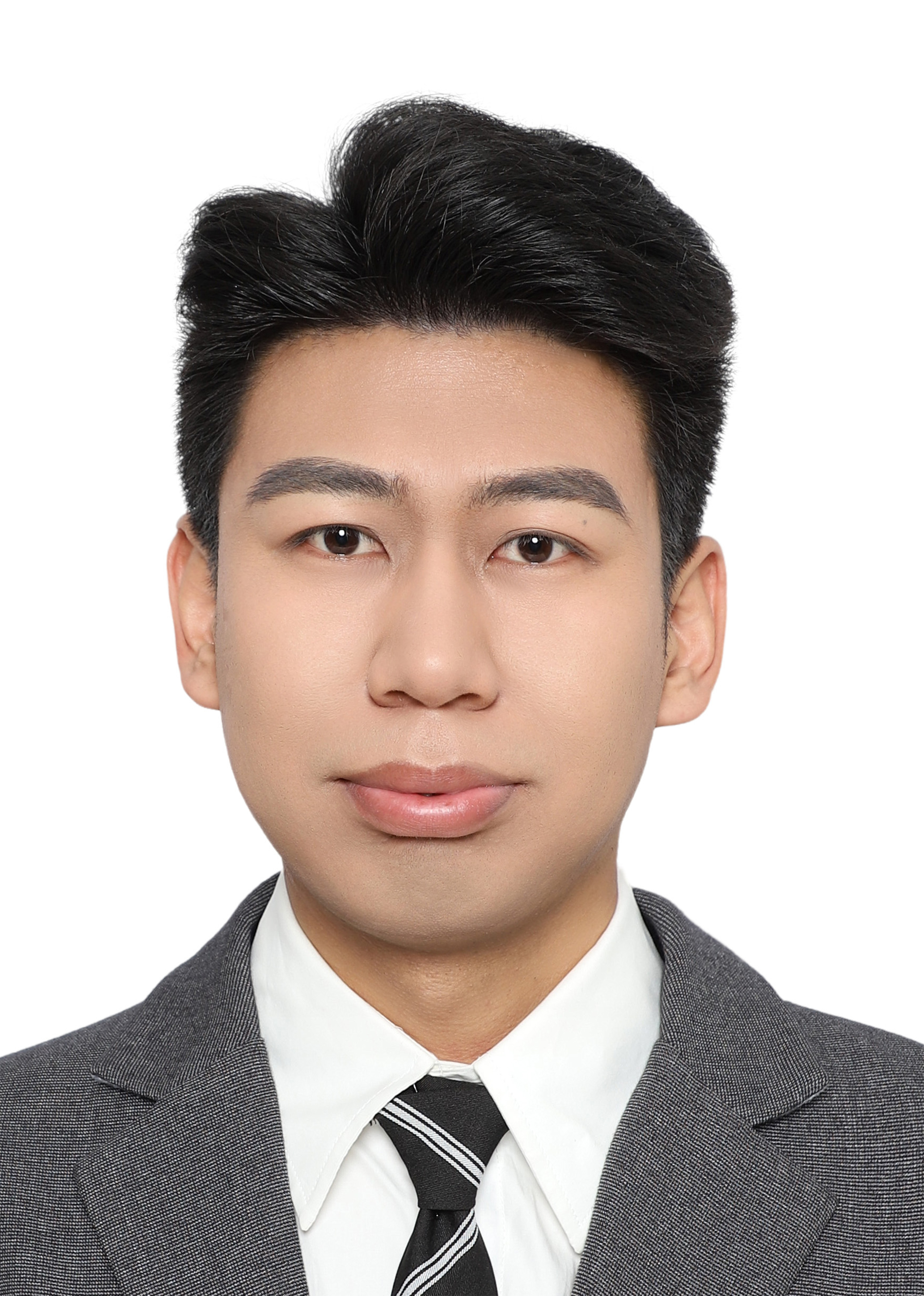}}]{Zhenwu Shi} received the M.S. degree in Educational Information Technology from the National Engineering Research Center for E-Learning, Central China Normal University, in 2020. He is currently pursuing the Ph.D. degree in Intelligent Education at the Shanghai Institute of Intelligent Education, East China Normal University, Shanghai, China. His research interests include computer vision, human pose estimation, motion editing, and AI in education.
\end{IEEEbiography}

\vspace{-2em}
\begin{IEEEbiography}[{\includegraphics[width=1in,height=1.25in,clip,keepaspectratio]{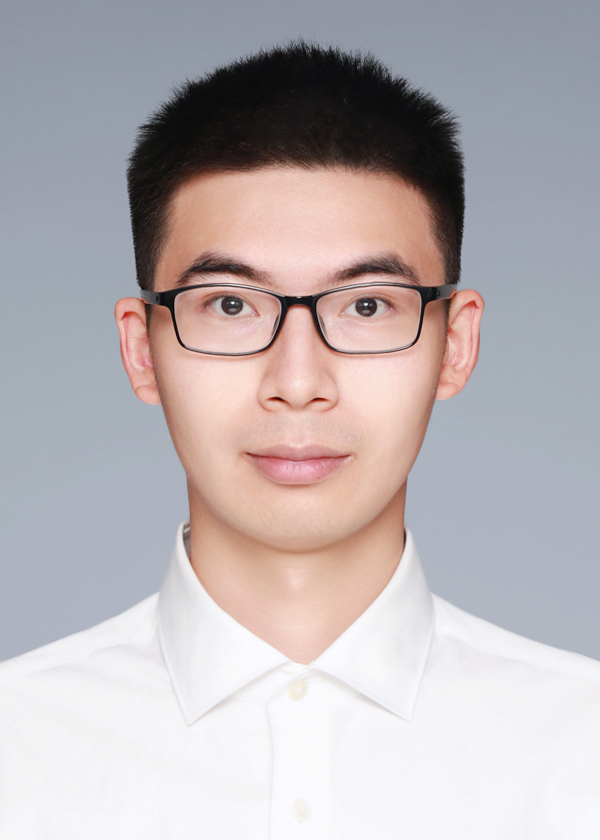}}]
{Jingyu Gong} received his Ph.D. from the Department of Computer Science and Technologies, Shanghai Jiao Tong University. He is now an Associate Research Professor with the School of Computer Science and Technology, East China Normal University, China. His research interests cover image processing and computer vision. He has published 15 papers in major international journals and conferences including the IEEE TPAMI, IEEE TVCG, CVPR, ECCV, AAAI, IJCAI, ACM MM, etc.
\end{IEEEbiography}

\vspace{-2em}
\begin{IEEEbiography}[{\includegraphics[width=1in,height=1.25in,clip,keepaspectratio]{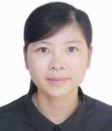}}]
{Jiao Xie} received a Ph.D. degree from Xiamen University, Fujian, China, in 2023. She is currently a PostDoc. at East China Normal University. Her research interests include Computer Vision and Deep Learning.
\end{IEEEbiography}

\vspace{-2em}
\begin{IEEEbiography}[{\includegraphics[width=1in,height=1.25in,clip,keepaspectratio]{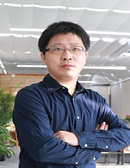}}]
{Zhou Yu}, School of Statistics, Key Laboratory of Advanced Theory and Application in Statistics and Data Science, Ministry of Education (KLATASDS-MOE), East China Normal University, Shanghai, China. 
Zhou Yu received the B.Sc. degree from the Hefei University of Technology in 2004 and the Ph.D. degree from East China Normal University, China, in 2010.
He is currently a Professor and the Ph.D. Supervisor with East China Normal University, and the Dean of the School of Statistics, ECNU. He has published more than 40 articles in well-known statistics journals, such as Annals of Statistics, Biometrika, and Journal of the American Statistical Association. His current research interests include statistical analysis of high-dimensional data and statistical machine learning.
\end{IEEEbiography}

\vspace{-2em}
\begin{IEEEbiography}[{\includegraphics[width=1in,height=1.25in,clip,keepaspectratio]{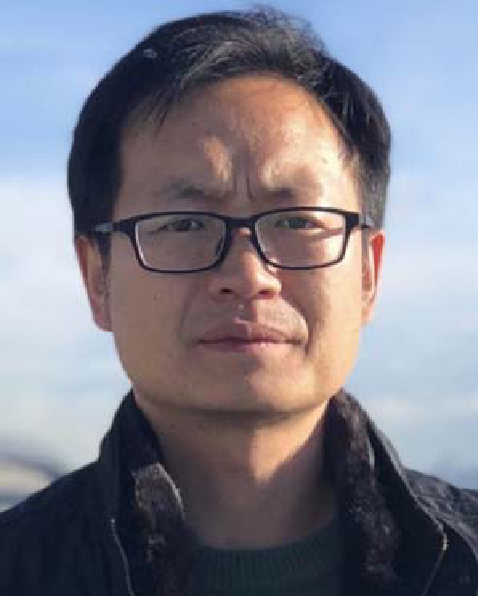}}]
{Baochang Zhang} received his Ph.D. degree from Harbin Institute of Technology. He is a professor at Beihang University, and Zhongguancun National Laboratory and an academic advisor at Baidu’s Deep Learning Laboratory. His research interests include model compression and acceleration, as well as efficient object recognition.
\end{IEEEbiography}

\vspace{-3em}
\begin{IEEEbiography}[{\includegraphics[width=1in,height=1.25in,clip,keepaspectratio]{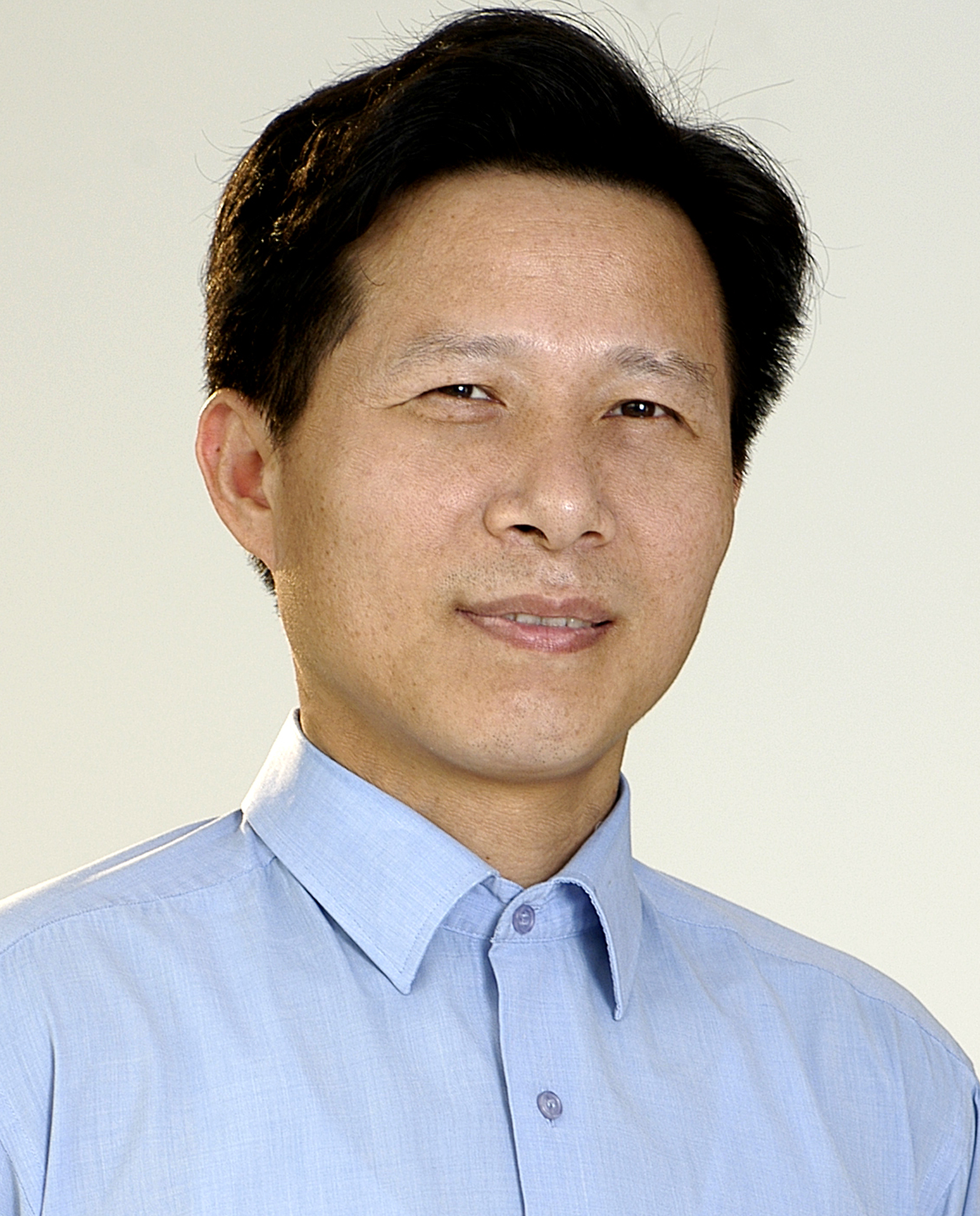}}]{Lizhuang Ma} received his B.S. and Ph.D. degrees from Zhejiang University, China, in 1985 and 1991, respectively. He is now a Distinguished Professor, at the Department of Computer Science and Engineering, Shanghai Jiao Tong University, China, and the School of Computer Science and Technology, East China Normal University, China. He was a Visiting Professor at the Frounhofer IGD, Darmstadt, Germany in 1998, and a Visiting Professor at the Center for Advanced Media Technology, Nanyang Technological University, Singapore from 1999 to 2000. His research interests include computer vision, computer aided geometric design, computer graphics, scientific data visualization, computer animation, digital media technology, and theory and applications for computer graphics, CAD/CAM. He serves as the reviewer of IEEE TPAMI, IEEE TIP, IEEE TMM, CVPR, AAAI etc.
\end{IEEEbiography}

\begin{IEEEbiography}[{\includegraphics[width=1in,height=1.25in,clip,keepaspectratio]{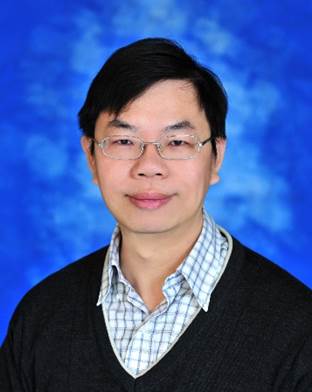}}]
{Chia-Wen Lin} received his Ph.D. degree from the Department of Electrical Engineering, National Tsing Hua University (EE/NTHU), Hsinchu, Taiwan, in January 2000. He is a professor with EE/NTHU since August 2007. Prior to joining EE/NTHU in 2007, he worked for the Department of Computer Science and Information Engineering, National Chung Cheng University (CSIE/CCU), Chiayi, Taiwan from August 2000 to July 2007. He was with the Information and Communications Research Laboratories, Industrial Technology Research Institute (ICL/ITRI), Hsinchu, Taiwan, during 1992~2000, where his final post was Section Manager. He served as a visiting scholar at the Information Processing Laboratory, Department of Electrical Engineering, University of Washington, USA during April~August 2000, a visiting professor with Microsoft Research Asia, Beijing, China, during July-August 2002, with National Institute of Informatics, Tokyo, Japan from August 2015 to February 2016, and with Nagoya University during June~December 2019. His research interests include visual content analysis and processing, and video networking.

Dr. Lin was named \textbf{IEEE Fellow} for his contributions to multimedia coding and editing in 2018. He is serving on the Board-of-Governors (2022~2024) and Fellow Evaluation Committee (2021~2023) of IEEE Circuits and Systems Society (CASS). He was a Distinguished Lecturer of IEEE CASS (2018~2019). He also served as President of the Chinese Image Processing and Pattern Recognition Association, Taiwan (2019~2020). His paper won the Young Investigator Award of SPIE VCIP 2005 and Best Paper Award of IEEE VCIP 2015. He was a recipient of the Distinguished Research Award (2023) and the Ta-You Wu Memorial Award (2006) both presented by National Science \& Technology Council (NSTC), Taiwan. He has served on the editorial board of IEEE Transactions on Image Processing, IEEE Transactions on Circuits and Systems for Video Technology, IEEE Transactions on Multimedia, IEEE Multimedia Magazine, Journal of Visual Communication and Image Representation, and Signal Processing: Image Communication. He also served as a Guest Editor of five special issues for IEEE Journal of Selected Topics in Signal Processing, IEEE Transactions on Multimedia,  EURASIP Journal on Advances in Signal Processing, and Journal of Visual Communication and Image Representation, respectively. He was Chair of the Multimedia Systems and Applications Technical Committee of IEEE Circuits and Systems Society (2013~2015).  He served as Steering Committee Chair of IEEE ICME (2020~2021), TPC Co-Chair of IEEE ICIP 2019 and ICME 2010, General Co-Chair of IEEE VCIP 2018 and PCS 2024, and Special Session Co-Chair of IEEE ICME 2009 and ICME 2018.
\end{IEEEbiography}

\vfill

\end{document}